\documentclass{article}

\usepackage{PRIMEarxiv}

\usepackage[utf8]{inputenc} % allow utf-8 input
\usepackage[T1]{fontenc}    % use 8-bit T1 fonts
\usepackage{hyperref}       % hyperlinks
\usepackage{url}            % simple URL typesetting
\usepackage{booktabs}       % professional-quality tables
\usepackage{amsfonts}       % blackboard math symbols
\usepackage{nicefrac}       % compact symbols for 1/2, etc.
\usepackage{microtype}      % microtypography
\usepackage{lipsum}
\usepackage{fancyhdr}       % header
\usepackage{graphicx}       % graphics
\graphicspath{{media/}}     % organize your images and other figures under media/ folder
\usepackage{natbib}
\usepackage{doi}
\usepackage{amsmath} 
\usepackage{amssymb}
\usepackage{xcolor}
\usepackage{subcaption}
\usepackage{tabularx}
\usepackage{multirow}
\usepackage{algorithm}
\usepackage{algpseudocode}
\usepackage{comment}
\usepackage{float}
\usepackage{placeins}

\title{Curriculum Learning-Driven PIELMs for Fluid Flow Simulations
}

\author{
  Vikas Dwivedi\thanks{Corresponding Author} \\
  CREATIS Biomedical Imaging Laboratory \\
  INSA, CNRS UMR 5220, Inserm, Universit´e Lyon 1 \\
  Lyon 69621\\
  \texttt{vikas.dwivedi@creatis.insa-lyon.fr} \\
   \And
  Bruno Sixou \\
  CREATIS Biomedical Imaging Laboratory \\
  INSA, CNRS UMR 5220, Inserm, Universit´e Lyon 1 \\
  Lyon 69621\\
  \texttt{bruno.sixou@insa-lyon.fr} \\
     \And
  Monica Sigovan \\
  CREATIS Biomedical Imaging Laboratory \\
  INSA, CNRS UMR 5220, Inserm, Universit´e Lyon 1 \\
  Lyon 69621\\
  \texttt{monica.sigovan@insa-lyon.fr} \\
}

\begin{document}
\maketitle

\begin{abstract}
This paper presents two novel, physics-informed extreme learning machine (PIELM)-based algorithms for solving steady and unsteady nonlinear partial differential equations (PDEs) related to fluid flow. Although single-hidden-layer PIELMs outperform deep physics-informed neural networks (PINNs) in speed and accuracy for linear and quasilinear PDEs, their extension to nonlinear problems remains challenging. To address this, we introduce a curriculum learning strategy that reformulates nonlinear PDEs as a sequence of increasingly complex quasilinear PDEs. Additionally, our approach enables a physically interpretable initialization of network parameters by leveraging Radial Basis Functions (RBFs). The performance of the proposed algorithms is validated on two benchmark incompressible flow problems: the viscous Burgers equation and lid-driven cavity flow. To the best of our knowledge, this is the first work to extend PIELM to solving Burgers' shock solution as well as lid-driven cavity flow up to a Reynolds number of 100.  As a practical application, we employ PIELM to predict blood flow in a stenotic vessel. The results confirm that PIELM efficiently handles nonlinear PDEs, positioning it as a promising alternative to PINNs for both linear and nonlinear PDEs.
\end{abstract}


%-----------------------------------------------------------------------------------------------------------------------------------------------%
\section{Introduction}\label{Sec:Introduction}
Nonlinear partial differential equations (PDEs) play a fundamental role in fluid mechanics, governing complex phenomena in aerodynamics, weather prediction, and engineering simulations. The Navier-Stokes equations, which describe the motion of viscous fluids, are central to understanding a wide range of flow regimes. Low-speed (weakly nonlinear) flows, such as creeping flow in microfluidic devices, blood flow in large arteries, and laminar boundary layers over airfoils, are often dominated by viscous effects and exhibit smooth solutions. In contrast, high-speed (strongly nonlinear) flows, including shock waves in supersonic jets, turbulence in atmospheric dynamics, and combustion-driven flows, involve sharp gradients, discontinuities, and chaotic behavior, making them significantly harder to solve.

Traditionally, computational fluid dynamics (CFD) is based on numerical techniques such as the finite element method\citep{Reddy2019} (FEM), the finite volume method\citep{versteeg2007introduction} (FVM) and spectral methods\citep{canuto2007spectral}. While these methods provide high accuracy, they become computationally expensive, especially for high-dimensional or highly nonlinear problems that require very fine spatial and temporal resolution.

To address these challenges, machine learning (ML)-based solvers have recently emerged as a promising alternative. In particular, Physics-Informed Neural Networks\citep{RAISSI2019686} (PINNs) and their variants\citep{Karniadakis2021} leverage data-driven learning while incorporating physical constraints to approximate PDE solutions efficiently. Despite their flexibility, PINNs often suffer from slow convergence, particularly for high-frequency solutions, and incur significant computational costs\citep{zhang2024physics}. 

To improve computational efficiency, Dwivedi and Srinivasan\citep{DWIVEDI202096} introduced single-hidden-layer Physics-Informed Extreme Learning Machines (PIELMs) and their distributed variant\citep{DWIVEDI2021299} (DPIELM). Unlike PINNs, which optimize all network parameters through iterative backpropagation, PIELMs are linear in the outer-layer weights, allowing single-pass optimization. This significantly accelerates computations for linear PDEs\citep{DWIVEDI202096}\citep{DONG2021114129}. PIELMs offer advantages in speed, but they also face two key challenges:
\begin{itemize}
\item \textbf{Limited applicability to nonlinear PDEs}: Although PIELMs outperform PINNs for linear problems, their direct extension to nonlinear PDEs remains difficult, limiting their broader applicability. 
\item \textbf{Lack of a systematic parameter initialization strategy}: PIELMs typically assign hidden-layer coefficients randomly within a user-defined range, making accuracy highly sensitive to these choices. However, there is no established method to select these parameters in a systematic or physically interpretable manner.
\end{itemize}   

Some prior efforts have attempted to address these limitations, such as iterative nonlinear least-squares methods\citep{DONG2021114129} for handling nonlinear PDEs and differential evolution-based algorithms\citep{DONG2022111290} for optimizing initialization. However, these methods are mathematically complex, lack physical interpretability, and have been validated only on weakly nonlinear PDEs. Although a recent study\citep{CALABRO2021114188} introduced an interpretable parameter initialization method for ELM, its validation was limited to simple, steady-state, linear 1D advection-diffusion equations.

To overcome these limitations, this work proposes a curriculum learning\citep{bengio2009,NEURIPS2021_df438e52} strategy, reformulating nonlinear PDEs as a sequence of quasi-linear approximations, allowing PIELMs to gradually adapt to increasing complexity. Recently, SPINNs \citep{RAMABATHIRAN2021110600} have shown promising results in solving nonlinear PDEs by employing kernel-based neural networks with Radial Basis Functions (RBFs), offering a more interpretable formulation of the PINN hypothesis. Building on this idea, we integrate this RBF-based approach into the PIELM framework to enhance its physical interpretability.

These novelties enable classical PIELMs to efficiently tackle nonlinear fluid flows, bridging the gap between classical numerical solvers and machine learning-based approaches while maintaining computational efficiency. 

%-----------------------------------------------------------------------------------------------------%
\paragraph{\textbf{Organization:}}
The structure of this paper is outlined as follows: Section \ref{Sec:Recap} briefly recaps PINNs, PIELMs, RBF methods for PDEs, and curriculum learning. Section \ref{Sec:Methodology} presents the proposed methodology. Section \ref{Sec:Results and Discussion} discusses the results on Burgers' equation, lid-driven cavity flow and stenotic flow, and Section \ref{Sec:Conclusion} concludes with key findings and future directions.

%-----------------------------------------------------------------------------------------------------%
%                                      Recap                                                          %
%-----------------------------------------------------------------------------------------------------%
\section{Background}\label{Sec:Recap}
%-----------------------------------------------------------------------------------------------------%
\begin{figure}[ht]
 \centerline{\includegraphics[width=0.7\textwidth,height=0.5\textheight]{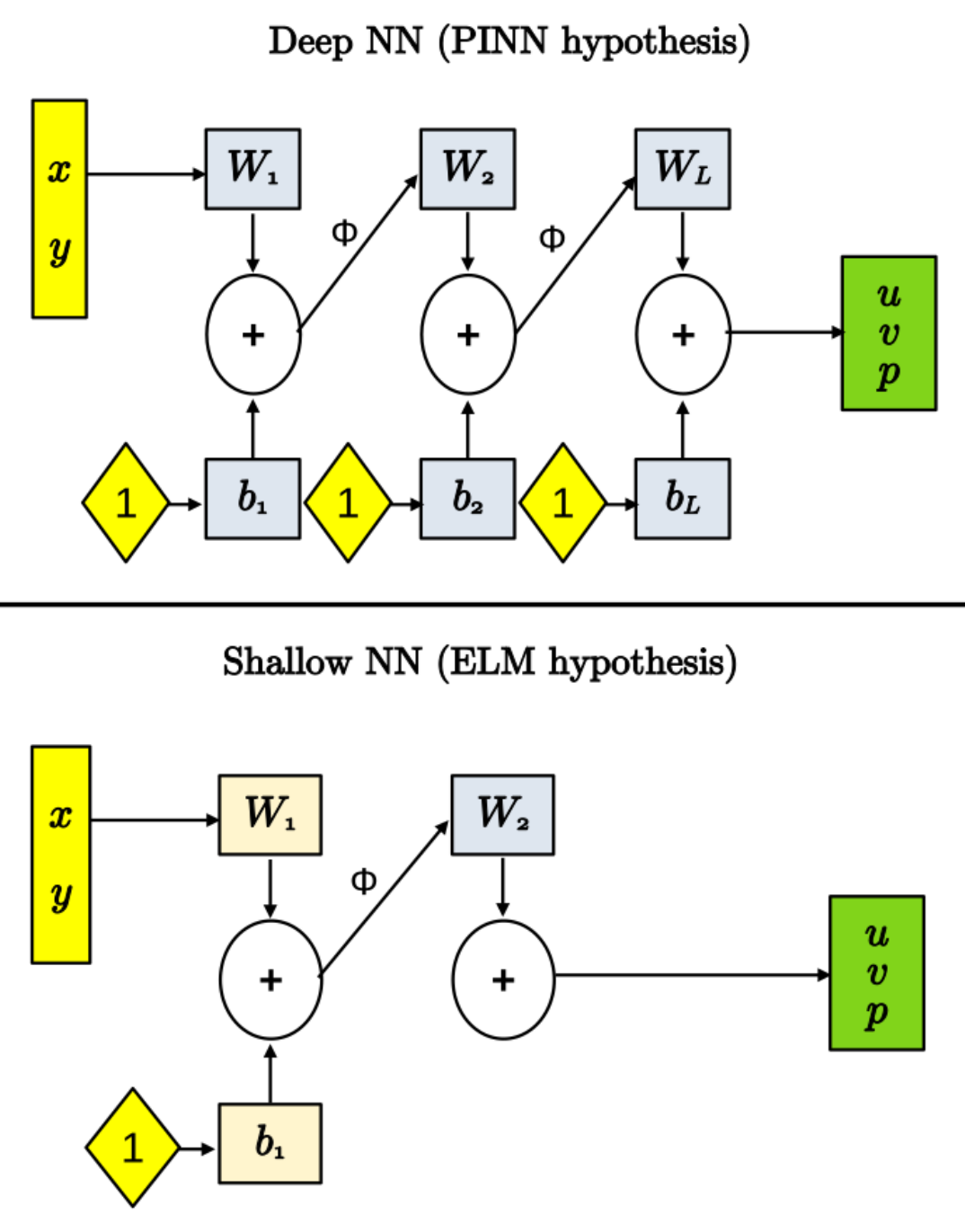}}
  \caption{Comparison of Network Architectures: PINN vs. PIELM. The figure illustrates the differences between deep PINNs (where all weights are trainable) and single-layer PIELMs (where input weights remain fixed). Yellow and green blocks indicate the input (spatial variables) and output (flow variables) of the networks. Network parameters, i.e. weights and biases, are denoted as $W_{i}$ and $b_{i}$. Trainable parameters are shown in gray, while the fixed input layer parameters of PIELM are highlighted in light orange.}
  \label{fig: Hypothesis}
\end{figure}
\begin{table}
  \centering
  \begin{tabular}{|c|c|c|}
  \hline
  \textbf{Property} & \textbf{PINN } & \textbf{PIELM}  \\
  \hline
  Hypothesis & Nonlinear in hidden layer weights & Linear in outer layer weights \\
  \hline
  Meshing & Not required  & Not required  \\  
  \hline
  Unknowns & Weights and biases of all layers & Weights of only output layer  \\
  \hline
  PDE Information & Differentiable total PDE residual & Linear system of PDE residuals \\
  \hline
  Optimization  & Gradient descent & Pseudo/Matrix inverse  \\
  \hline
  Applicability  & Linear and nonlinear PDEs & Linear and quasi-linear PDEs  \\  
  \hline
  Speed  & Low as compared to PIELM & High as compared to PINN  \\    
  \hline
\end{tabular}
\caption{Comparison of PINN and PIELM}
\label{Tab:comparison}
\end{table}
%-----------------------------------------------------------------------------------------------------%
\paragraph{\textbf{PINNs}} In a typical Physics-Informed Neural Network (PINN) framework, a deep neural network approximates the solution of PDEs. Randomly distributed collocation points within the computational domain, along with boundary points, serve as the training dataset. At these points, penalties are introduced for deviations from the governing PDE and boundary conditions (BCs). The mean squared error of these penalties, representing the residuals of the PDE and BCs, defines a physics-based loss function. This loss is minimized using gradient descent-based optimization algorithms and backpropagation. The core idea behind PINNs and their variants is to reformulate the task of solving PDEs as an optimization problem driven by a physics-informed loss function.

\paragraph{\textbf{PIELMs}}Physics-Informed Extreme Learning Machine (PIELM) differs from PINN in two key aspects: (A) network architecture and (B) cost function minimization. Unlike the deep multilayer architecture of PINNs, PIELM employs a single-hidden-layer design. A distinctive feature of PIELM is that its input layer weights are randomly initialized and remain fixed during training. With the first layer weights held constant, the PDE solution becomes linear with respect to the output layer weights, making the approximation process inherently linear. Instead of minimizing a physics-informed loss function through computationally intensive gradient descent and backpropagation techniques, PIELM formulates a system of linear residual equations, which can be efficiently solved using matrix inversion. Figure \ref{fig: Hypothesis} compares the PINN and PIELM hypotheses for solving the Navier-Stokes equations. Table \ref{Tab:comparison} provides a detailed comparison of their key properties. 

\paragraph{\textbf{RBF Methods for PDEs}}Radial Basis Function (RBF) kernels \citep{buhmann2000radial} are popular for solving PDEs due to their flexibility and ability to model complex, multidimensional problems. These real-valued functions depend on the distance between points, making them ideal for interpolating scattered data. Common RBFs, such as Gaussian, multi-quadric, and inverse multi-quadric, are parameterized to control smoothness and influence radius. In the context of PDEs, RBFs approximate solutions by constructing a weighted sum of basis functions centered at collocation points, which satisfy the governing equations and boundary conditions \citep{KANSA1990127, KANSA1990147}. This approach is computationally efficient, as it avoids iterative optimization, but the performance is heavily dependent on on the careful selection of kernel parameters \citep{Fasshauer2007} and collocation points.

\paragraph{\textbf{Curriculum Learning}} Curriculum learning\citep{bengio2009, bengio2009learning} is a machine learning strategy where a model is trained by presenting examples in a sequence of increasing difficulty, similar to how humans learn. The idea is to start with simpler, easier-to-learn tasks and gradually progress to more complex ones. This approach helps the model to converge more efficiently by providing a structured learning path, avoiding early confusion with overly difficult examples\citep{elman1993learning}. Curriculum learning has been successfully applied in various fields, such as reinforcement learning\citep{parisotto2017neural}, natural language processing\citep{campos2021curriculumlearninglanguagemodeling}, and computer vision\citep{wang2024EfficientTrain_pp}, where it has been shown to improve the training stability, convergence rate, and generalization performance of models.

%-----------------------------------------------------------------------------------------------------%
%                                      Methods                                                        %
%-----------------------------------------------------------------------------------------------------%
\section{Methods}\label{Sec:Methodology}
%-----------------------------------------------------------------------------------------------------%
%\textcolor{blue}{ mention the dimension of the problem in each case, and define function for example u:$R \rightarrow ....., x \in R, \Omega ... $}
\subsection{Test Cases}
To evaluate the performance of PIELM, we have selected two standard nonlinear PDEs from fluid dynamics: (1) Burgers' equation and (2) Navier-Stokes equations. 

\subsubsection{Burgers' Equation}

Burgers' equation models nonlinear advection with linear diffusion, and is expressed as follows:
\begin{equation}
\frac{\partial u}{\partial t} + u \frac{\partial u}{\partial x} = \nu \frac{\partial^{2} u}{\partial x^{2}}, \quad (x, t) \in \Omega
\label{eqn:Burgers}
\end{equation}

where:

\begin{itemize}
    \item \( u: \Omega \rightarrow \mathbb{R} \) is the velocity field,
    \item \( x \in \mathbb{R} \) is the spatial coordinate,  
    \item \( t \in \mathbb{R} \) is time,  
    \item \( \nu \in \mathbb{R} \) is the viscosity coefficient,  
    \item \( \Omega \subset \mathbb{R}^2 \) represents the spatio-temporal computational domain (rectangle in $(x, t)$).
\end{itemize}

The equation describes the balance between the nonlinear advection term \( u \frac{\partial u}{\partial x} \) and the linear diffusion term \( \nu \frac{\partial^{2} u}{\partial x^{2}} \). For small values of \( \nu \), the solution exhibits sharp gradients, and as \( \nu \to 0 \), the solution may develop discontinuities, commonly referred to as shocks. These properties make Burgers' equation particularly useful for modeling abrupt changes in flow behavior.

\subsubsection{Navier-Stokes Equations}

The Navier-Stokes equations describe the motion of viscous fluid substances. We restrict here to steady-state solutions. The steady-state, incompressible form of these equations in two dimensions is expressed as:

\begin{equation}
\nabla \cdot \overrightarrow{V} = 0, \quad (x, y) \in \Omega
\label{eqn:cont}
\end{equation}

\begin{equation}
(\overrightarrow{V} \cdot \nabla) \overrightarrow{V} = -\nabla p + \frac{1}{Re} \nabla^2 \overrightarrow{V}, \quad (x, y) \in \Omega
\label{eqn:mom}
\end{equation}

where:

\begin{itemize}
    \item \( \overrightarrow{V} = \left( \begin{array}{c} u \\ v \end{array} \right): \Omega \rightarrow \mathbb{R}^2 \) is the velocity field,
    \item \( u, v \in \mathbb{R} \) are the velocity components in the \( x \)- and \( y \)-directions, respectively,  
    \item \( p: \Omega \rightarrow \mathbb{R} \) is the pressure field, 
    \item \( Re \in \mathbb{R} \) is the Reynolds number,  
    \item \( \nabla \) and \( \nabla^2 \) denote the gradient and Laplacian operators, respectively,  
    \item \( \Omega \subset \mathbb{R}^2 \) represents the 2D computational domain, and \( \partial\Omega \) is its boundary.
\end{itemize}

Equation \ref{eqn:cont} describes mass conservation, while Equation \ref{eqn:mom} represents momentum conservation in the horizontal and vertical directions. The left side of the momentum equation contains the inertial terms, while the right side contains the pressure term and the viscous term. The balance between the inertia and viscous forces determines the flow behavior. The Reynolds number (\( Re \)) characterizes the ratio of inertia to viscous forces: high values of \( Re \) correspond to turbulent flows, while low values indicate viscosity-dominated slow flows. Practically, the Reynolds number is defined as:

\[
Re = \frac{UL}{\nu}
\]

where \( L \) is the characteristic length, \( U \) is the characteristic velocity and \( \nu \) is the kinematic viscosity. The pressure term primarily ensures mass conservation.
%-----------------------------------------------------------------------------------------------------%
\subsection{Proposed Curriculum Learning for Nonlinear PDEs}
%-----------------------------------------------------------------------------------------------------%
%\textcolor{msc}{there is redundance here with what you said previouslt for CL so maybe reduce it a bit}
Curriculum learning\citep{bengio2009} follows a progression from simpler concepts to more complex ones. When applied to solving PDEs, this approach involves first solving simpler PDEs before moving on to more complex ones \citep{NEURIPS2021_df438e52}. PIELMs have shown promising results in solving linear and quasi-linear PDEs. However, their inherently linear formulation prevents their direct application to nonlinear problems. We introduce curriculum learning-based algorithms for solving both unsteady and steady nonlinear PDEs. While the ideas are general, we explain them in the context of our test cases: Burgers' equation for unsteady problems and the Navier-Stokes equations for steady cases.

\subsubsection{For Burgers' Equation}
The nonlinear advection term $u\frac{\partial u}{\partial x}$ in Burgers' equation\ref{eqn:Burgers} is the main modeling challenge for PIELM which is based on a linear representation of the solution. To address this, we approximate it using a quasi-linear formulation. However, the accuracy of this linearized model is generally limited to small deviations from the chosen linearization point. To overcome this limitation, rather than solving the PDE over the entire computational domain at once, we partition the domain into multiple time blocks and solve the linearized Burgers' equation sequentially using a predictor-corrector approach. 

Assume that we partition the computational domain \( \Omega = [0, T] \times [-1, 1] \) into \( B \) time blocks such that:

\[
\Omega = \bigcup_{i=0}^{B-1} \Omega_i, \quad \text{where} \quad \Omega_i = \left[ i \cdot dT, (i+1) \cdot dT \right] \times [-1, 1], \quad i = 0, 1, 2, \dots, B-1,
\]

with each block \( \Omega_i \) corresponding to the time interval \( \left[ i \cdot dT, (i+1) \cdot dT \right] \), where \( dT \) is the time step size, and \( T = B \cdot dT \) is the total time. In the first time block $(B=1)$, the linearization point is set by the initial condition. We first solve the predictor step, where the linearized Burgers' equation is solved using the initial condition as the reference velocity:
\begin{equation}
u_{pred}=u(x,t=0)
\end{equation}
\begin{equation}
\frac{\partial u}{\partial t}+u_{pred}\frac{\partial u}{\partial x}=\nu\frac{\partial^{2}u}{\partial x^{2}}
\end{equation}
Next, in the corrector step, we refine the solution by solving the linearized equation again, using the predictor solution ($\widehat{u}_{B=1}(x,t)$) as the updated reference:
\begin{equation}
u_{corr}=\widehat{u}_{B=1}(x,t)
\end{equation}
\begin{equation}
\frac{\partial u}{\partial t}+u_{corr}\frac{\partial u}{\partial x}=\nu\frac{\partial^{2}u}{\partial x^{2}}
\end{equation}
This iterative predictor-corrector process is applied across all time blocks. The full pseudocode for this procedure is presented in Algorithm \ref{algo:Burgers}.

\begin{algorithm}
\caption{Curriculum learning-based PIELM for Burgers' equation}
\begin{algorithmic}[1]
\State Initialize $B \gets 0$, $u(x,t_{init}) \gets F_{ic}(x)$
\State Approximate $uu_{x} \gets u(x,t_{init})u_{x}$
\State Solve Burgers equations with quasi-linear advection terms (predictor-step)
\State Approximate $uu_{x}$ using $u$ from predictor-step.
\State Solve Burgers equations with quasi-linear advection terms (corrector-step)
\State $B \gets B + 1$
\State $u(x,t_{init}))_{B} \gets (u(x,t_{end}))_{B-1}$
\While{$B \leq B_{target}$}
    \State Approximate $uu_{x} \gets u(x,t_{init})u_{x}$
    \State Solve Burgers equations with quasi-linear advection terms (predictor-step)
    \State Approximate $uu_{x}$ using $u$ from predictor-step.
    \State Solve Burgers equations with quasi-linear advection terms (corrector-step)
    \State $B \gets B + 1$
    \State $u(x,t_{init}))_{B} \gets (u(x,t_{end}))_{B-1}$
\EndWhile
\end{algorithmic}
\label{algo:Burgers}
\end{algorithm}
%-----------------------------------------------------------------------------------------------------%
\subsubsection{For Navier-Stokes Equation}
Similar to Burgers' equation, the nonlinear advection term $\left(\overrightarrow{V}.\nabla\right)\overrightarrow{V}$ in the Navier-Stokes equations is the main problem for PIELM.
To address this, we adopt a quasi-linear approximation. However, unlike Burgers' equation, a steady problem cannot be decomposed into spatial blocks, nor do we have a predefined reference solution, such as an initial condition, for linearization. 

To resolve this, we initially set $\left(\overrightarrow{V}.\nabla\right)\overrightarrow{V}=\overrightarrow{0}$ and use the Stokes flow solution as the starting linearization point. The PIELM solution for Stokes flow serves as the initial condition, and for the quasi-linear approximation. Instead of progressing through time blocks, we advance incrementally in Reynolds number $(Re)$. The complete pseudocode for this approach is provided in Algorithm \ref{algo:NSE}.

\begin{algorithm}
\caption{Curriculum learning-based PIELM for Navier-Stokes equations}
\begin{algorithmic}[1]
\State Initialize $Re \gets 0$, $\delta \gets \delta_{user}$
\State Solve Navier-Stokes equations ignoring advection terms (Stokes flow)
\State Approximate $(\mathbf{V} \cdot \nabla) \mathbf{V} \gets (\mathbf{V}_{Stokes} \cdot \nabla) \mathbf{V}$
\State Solve Navier-Stokes equations including quasi-linear advection terms
\State $Re \gets Re + \delta$
\While{$Re \leq Re_{target}$}
    \State Approximate $(\mathbf{V} \cdot \nabla) \mathbf{V} \gets (\mathbf{V}_{Re-\delta} \cdot \nabla) \mathbf{V}$
    \State Solve Navier-Stokes equations with quasi-linear advection terms
    \State $Re \gets Re + \delta$
\EndWhile
\end{algorithmic}
\label{algo:NSE}
\end{algorithm}
%-----------------------------------------------------------------------------------------------------%
\subsection{PIELM Formulation}
%-----------------------------------------------------------------------------------------------------%
We present a formulation based on the Navier-Stokes equations. The computational domain is two-dimensional for both the Burgers equations and the Navier-Stokes equations, with $(x,t)$ coordinates for Burgers and $(x,y)$ for Navier-Stokes. Consequently, the interpretation of the PIELM input parameters will remain unchanged. However, while the PIELM output for the Burgers equation is a scalar $u$, it is a vector $(u,v,p)$ for the Navier-Stokes equations. Therefore, the reader can derive the formulation for the Burgers equation as a special case of the Navier-Stokes equations.

\subsubsection{Hypothesis}

Consider two column vectors, \( \overrightarrow{\alpha^{*}} \) and \( \overrightarrow{\beta^{*}} \), which represent the \( x \)- and \( y \)-coordinates of the centers of the RBF kernel, respectively. In addition, we define two scaling vectors, \( \overrightarrow{m} \) and \( \overrightarrow{n} \), which scale the coordinates along the \( x \)- and \( y \)-directions, respectively. Now, we define new vectors \( \overrightarrow{\alpha} \) and \( \overrightarrow{\beta} \) by performing element-wise multiplication of the scaling vectors \( \overrightarrow{m} \) and \( \overrightarrow{n} \) with \( \overrightarrow{\alpha^{*}} \) and \( \overrightarrow{\beta^{*}} \), respectively. Specifically, the vectors \( \overrightarrow{\alpha} \) and \( \overrightarrow{\beta} \) are given by:

\[
\overrightarrow{\alpha} = \overrightarrow{m} \odot \overrightarrow{\alpha^{*}}, \quad \overrightarrow{\beta} = \overrightarrow{n} \odot \overrightarrow{\beta^{*}},
\]

where \( \odot \) denotes element-wise multiplication (also known as the Hadamard product). Then, the set of vectors $(\overrightarrow{\alpha},\overrightarrow{\beta},\overrightarrow{m},\overrightarrow{n})$ constitute the parameters of the input layer of ELM. In the ELM architecture, the input layer parameters are user-selected and not trainable. Consequently, the output is expressed as a linear combination of the outer layer weights.

For 2D steady Navier-Stokes equations, the hypothesis of proposed PIELM (refer Figure\ref{fig: Hypothesis}) is given by

\begin{equation}
\hat{u}(x,y)=\sum_{k=1}^{k=N^{*}}\phi(z_{k}(x,y))c_{u,k}
\end{equation}
\begin{equation}
\hat{v}(x,y)=\sum_{k=1}^{k=N^{*}}\phi(z_{k}(x,y))c_{v,k}
\end{equation}
\begin{equation}
\hat{p}(x,y)=\sum_{k=1}^{k=N^{*}}\phi(z_{k}(x,y))c_{p,k}
\end{equation}
where $\overrightarrow{c}=\left(\begin{array}{c}
\overrightarrow{c_{u}}\\
\overrightarrow{c_{v}}\\
\overrightarrow{c_{p}}
\end{array}\right)$ denotes outer layer weights ($W_{2}$ of PIELM hypothesis in Figure\ref{fig: Hypothesis}), $N^{*}$denotes the number of neurons
in the hidden layer, $\phi$ is the radial basis function defined
by $\phi(z)=e^{-z^{2}}$, and $z_{k}(x,y)$ is given by
\begin{equation}
z_{k}(x,y)=\sqrt{(m_{k}x+\alpha_{k})^{2}+(n_{k}y+\beta_{k})^{2}}
\label{eqn:z_k}
\end{equation}where $(m_{k},n_{k},\alpha_{k},\beta_{k})$ denote elements of $(\overrightarrow{m},\overrightarrow{n},\overrightarrow{\alpha},\overrightarrow{\beta})$. The input layer parameters for $(u,v,p)$ are kept same to respect the pressure-velocity coupling.

Suppose that we define $\xi=-z^{2}$. Then the expressions for the first derivative terms are as follows: 
\begin{equation}
\frac{\partial\hat{u}}{\partial x}=\sum_{k=1}^{k=N^{*}}\frac{\partial\phi}{\partial\xi_{k}}\frac{\partial\xi_{k}}{\partial x}c_{u,k}=\sum_{k=1}^{k=N^{*}}-2e^{\xi_{k}}m_{k}(m_{k}x+\alpha_{k})c_{u,k}
\end{equation}

\begin{equation}
\frac{\partial\hat{u}}{\partial y}=\sum_{k=1}^{k=N^{*}}\frac{\partial\phi}{\partial\xi_{k}}\frac{\partial\xi_{k}}{\partial y}c_{u,k}=\sum_{k=1}^{k=N^{*}}-2e^{\xi_{k}}n_{k}(n_{k}y+\beta_{k})c_{u,k}
\end{equation} 

Similarly, we can derive expressions for the remaining first derivative terms $\left(\frac{\partial v}{\partial x},\frac{\partial v}{\partial y}\right)$ and $\left(\frac{\partial p}{\partial x},\frac{\partial p}{\partial y}\right)$ by multiplying with corresponding outer layer weights for $v$ and $p$ respectively. 

The second derivative terms are given as follows:
\begin{equation}
\frac{\partial^{2}\hat{u}}{\partial x^{2}}=\sum_{k=1}^{k=N^{*}}-2e^{\xi_{k}}m_{k}^{2}\{1-2(m_{k}x+\alpha_{k})^{2}\}c_{u,k}
\end{equation}

\begin{equation}
\frac{\partial^{2}\hat{u}}{\partial y^{2}}=\sum_{k=1}^{k=N^{*}}-2e^{\xi_{k}}n_{k}^{2}\{1-2(n_{k}y+\beta_{k})^{2}\}c_{u,k}
\end{equation}
The expressions for $\left(\frac{\partial^{2}v}{\partial x^{2}},\frac{\partial^{2}v}{\partial y^{2}}\right)$
can be derived by multiplying with corresponding outer layer weights
for $v$.
\subsubsection{Interpretation of PIELM Parameters }
The proposed formulation, inspired by SPINN \citep{RAMABATHIRAN2021110600}, enables a physically interpretable initialization of network parameters by leveraging the properties of Radial Basis Functions (RBFs). Unlike random initialization methods, we directly associate the input layer parameters of the PIELM with meaningful physical quantities—specifically, the locations and characteristic scales of Gaussian functions.

The 2D Gaussian function, parameterized by mean and standard deviation vectors, is given by:
\begin{equation}
G\left(x,y;\alpha^{*},\beta^{*},\sigma_{x},\sigma_{y}\right)=e^{-\left(\frac{(x-\alpha^{*})^{2}}{2\sigma_{x}^{2}}+\frac{(y-\beta^{*})^{2}}{2\sigma_{y}^{2}}\right)}.
\end{equation}
On comparing this to the equation \ref{eqn:z_k} , we can deduce that the input layer parameters of the ELM $(\overrightarrow{\alpha},\overrightarrow{\beta},\overrightarrow{m},\overrightarrow{n})$ correspond to 
\begin{equation}
\left(-\frac{1}{\sqrt{2}\overrightarrow{\sigma_{x}}}\overrightarrow{\alpha^{*}},-\frac{1}{\sqrt{2}\overrightarrow{\sigma_{y}}}\overrightarrow{\beta^{*}},\frac{1}{\sqrt{2}\overrightarrow{\sigma_{x}}},\frac{1}{\sqrt{2}\overrightarrow{\sigma_{y}}}\right).
\end{equation}
Here, $\left(\overrightarrow{\alpha^{*}},\overrightarrow{\beta^{*}}\right)$ represent the spatial locations of the RBF centers, which are physically meaningful parameters that should be sampled from within the computational domain $\Omega$. Likewise, $\left(\overrightarrow{m},\overrightarrow{n}\right)$ represents the reciprocal of $\left(\overrightarrow{\sigma}_{x},\overrightarrow{\sigma}_{y}\right)$, which controls the spread of the activation functions. A structured flowchart outlining the process for selecting input layer parameters is presented below: 

%------------------------------------------%
\begin{enumerate}
\item {Start}
\item {Define Computational Domain ($\Omega$)}
\begin{itemize}
\item Identify the spatial boundaries and physical constraints of the problem.
\end{itemize}
\item {Determine RBF Centers ($\alpha^{*}$, $\beta^{*}$)}
\begin{itemize}
\item Sample spatial locations of RBF centers within $\Omega$.
\item Ensure distribution aligns with physical characteristics of the domain (e.g., more centers near walls).
\end{itemize}
\item {Define Characteristic Lengths ($L_c$)}
\begin{itemize}
\item Establish relevant length scales for the problem.
\end{itemize}
\item {Determine Standard Deviations ($\sigma_x, \sigma_y$)}
   \begin{itemize}
        \item Adjust based on sharp gradients:
        \begin{itemize}
            \item For Burgers' equation: Concentrate sharp RBF kernels (smaller $\sigma$) near steep gradients.
            \item For lid-driven cavity or stenotic flow: Concentrate sharp RBF kernels near walls or constriction.
        \end{itemize}
    \end{itemize}
\item {Compute Input Layer Parameters}
\begin{itemize}
\item $\alpha = - \frac{1}{\sqrt{2}\sigma_x} \alpha^{*}$
\item $\beta = - \frac{1}{\sqrt{2}\sigma_y} \beta^{*}$
\item $m = \frac{1}{\sqrt{2}\sigma_x}$
\item $n = \frac{1}{\sqrt{2}\sigma_y}$
\end{itemize}
\item {Initialize ELM with Computed Parameters}
\item {End}
\end{enumerate}

For specific cases, such as lid-driven cavity flow, a single characteristic length is sufficient, e.g., the distance between the cavity center and its corner. However, in more complex geometries, multiple length scales may be needed. For example, in a stenotic blood flow simulation, one length scale could be the inlet/outlet diameter, while another could represent the constriction diameter.

The proposed strategy bridges the gap between physics and machine learning, ensuring that the network's initial parameters have a well-grounded physical interpretation.
%------------------------------------------%
\subsubsection{Residuals}
\begin{enumerate}
\item \textit{PDE residual}: For each collocation point within $\Omega$, we calculate residuals from continuity, $x$ and $y$- momentum equations and set them equal to zero as follows:
\begin{equation}
\boldsymbol{M}\overrightarrow{c}=\overrightarrow{0}
\end{equation}
where $\boldsymbol{M}$ is a block matrix and $\overrightarrow{c}$ is outer layer weights matrix. Specifically,
\begin{equation}
\left[\begin{array}{ccc}
M_{u1} & M_{v1} & M_{p1}\\
M_{u2} & M_{v2} & M_{p2}\\
M_{u3} & M_{v3} & M_{p3}
\end{array}\right]\left(\begin{array}{c}
\overrightarrow{c_{u}}\\
\overrightarrow{c_{v}}\\
\overrightarrow{c_{p}}
\end{array}\right)=\left(\begin{array}{c}
0\\
0\\
0
\end{array}\right)
\end{equation}
Here the shape of individual blocks is $1 \times N^{*}$. The first row corresponds to the continuity equation. The second and third rows correspond to the $x$ and $y$- momentum equations, respectively. For $k=1,2,..,N^{*}$,
\begin{equation}
M_{u1}(1,k)=-2e^{\xi_{k}}m_{k}(m_{k}x+\alpha_{k})
\end{equation}
\begin{equation}
M_{v1}(1,k)=-2e^{\xi_{k}}n_{k}(n_{k}y+\beta_{k})
\end{equation}
\begin{equation}
M_{p1}(1,k)=0
\end{equation}
\begin{multline}
M_{u2}(1,k)=\hat{u}_{ref}M_{u1}(1,k)+\hat{v}_{ref}M_{v1}(1,k)\\
+\frac{2e^{\xi_{k}}}{Re}[m_{k}^{2}\{1-2(m_{k}x+\alpha_{k})^{2}\}+n_{k}^{2}\{1-2(n_{k}y+\beta_{k})^{2}\}]
\end{multline}
\begin{equation}
M_{v2}(1,k)=0
\end{equation}
\begin{equation}
M_{p2}(1,k)=M_{u1}(1,k)
\end{equation}
\begin{equation}
M_{u3}(1,k)=0
\end{equation}
\begin{equation}
M_{v3}(1,k)=M_{u2}(1,k)
\end{equation}
\begin{equation}
M_{p3}(1,k)=M_{v1}(1,k)
\end{equation}
The terms $\hat{u}_{ref}$ and $\hat{v}_{ref}$ represent the reference velocities used in the quasi-linear approximation of the nonlinear advection terms, as previously discussed.
\item \textit{Boundary condition residual}: For each boundary point in $\partial\Omega$, we can express the residual depending on the type of boundary condition
(Dirichlet, Neumann, or mixed). For example, the expression for no-slip condition is as follows:
\begin{equation}
\left[\begin{array}{ccc}
B_{u1} & B_{v1} & B_{p1}\\
B_{u2} & B_{v2} & B_{p2}
\end{array}\right]\left(\begin{array}{c}
\overrightarrow{c_{u}}\\
\overrightarrow{c_{v}}\\
\overrightarrow{c_{p}}
\end{array}\right)=\left(\begin{array}{c}
0\\
0
\end{array}\right)
\end{equation}
For $k=1,2,..,N^{*}$, 
\begin{equation}
B_{u1}(1,k)=e^{\xi_{k}}
\end{equation}
\begin{equation}
B_{v1}(1,k)=0
\end{equation}
\begin{equation}
B_{p1}(1,k)=0
\end{equation}
\begin{equation}
B_{u2}(1,k)=0
\end{equation}
\begin{equation}
M_{v2}(1,k)=B_{u1}(1,k)
\end{equation}
\begin{equation}
M_{p2}(1,k)=0
\end{equation}
Similarly, the expression for velocity inlet along horizontal direction can be written by modifiying the RHS term to $\left(\begin{array}{c}
U(y)\\
0
\end{array}\right).$
\end{enumerate}

\subsubsection{Optimization }

On assembling all the residuals, we typically get an over-determined system of linear equations of the form
\begin{equation}
\boldsymbol{A}\overrightarrow{c}=\overrightarrow{b}.
\end{equation}
%\textcolor{blue}{where A is the block matrix (M,B) ?}
The least square solution of $\overrightarrow{c}$ is found via least squares approach, i.e., 
\begin{equation}
\overrightarrow{c}=pinv(\boldsymbol{A})\overrightarrow{b}
\end{equation}
where $pinv$ refers to pseudo-inverse. This approach is much faster than back propagation-based gradient descent algorithms. 
%-----------------------------------------------------------------------------------------------------%
%                               Results and Discussion                                                %
%-----------------------------------------------------------------------------------------------------%
\section{Results and Discussion}\label{Sec:Results and Discussion}
In this section, we:  
\begin{enumerate}  
    \item Evaluate the performance of PIELM and PINN for solving the linear 2D Poisson equation, showing that PIELM surpasses PINN in both computational efficiency and accuracy.  
    \item Demonstrate, for the first time in the literature, that PIELM effectively captures shocks in the viscous Burgers' equation rapidly and with only half the training parameters required by PINN.  
    \item Extend PIELM to solve the steady Navier-Stokes equations for the lid-driven cavity benchmark problem, achieving accurate solutions for Reynolds numbers up to 100.  
    \item Apply PIELM to simulate stenotic blood flow at various Reynolds numbers as a real-world application.  
\end{enumerate}

We also emphasize that this paper is intended as a proof-of-concept study, and therefore, we have not performed a comprehensive analysis of hyperparameter selection. In particular, we do not present dedicated hyperparameter parameter selection studies on the influence of the number and distribution of sampling points, the placement and spread of RBF centers, or the effects of time step and Reynolds number increments. A thorough investigation into hyperparameter tuning and optimization is beyond the scope of this work and will be explored in future studies.

All the experiments are conducted in Matlab R2022b environment running in a 12th Gen Intel(R) Core(TM) i7-12700H, 2.30 GHz CPU and 16GB RAM Asus laptop.
\subsection{Poisson's Equation - PIELM versus PINN} 
%----------------------------------------------------------------------------------------------------------%
\begin{figure}[ht]
    \centering
    \begin{subfigure}{0.48\textwidth}
      \centering
      \includegraphics[width=\textwidth]{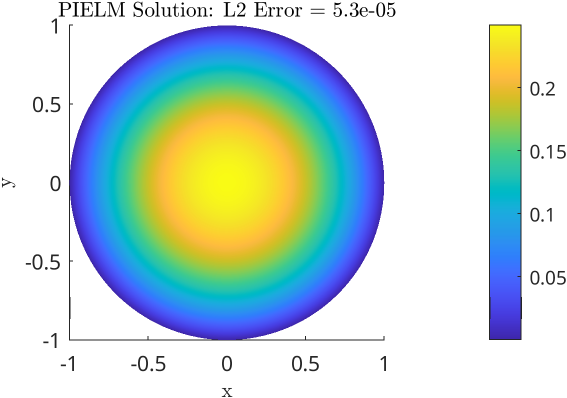}
      \caption{PIELM solution}
      \label{fig:pred_both}
    \end{subfigure}
    \begin{subfigure}{0.48\textwidth}
      \centering
      \includegraphics[width=\textwidth]{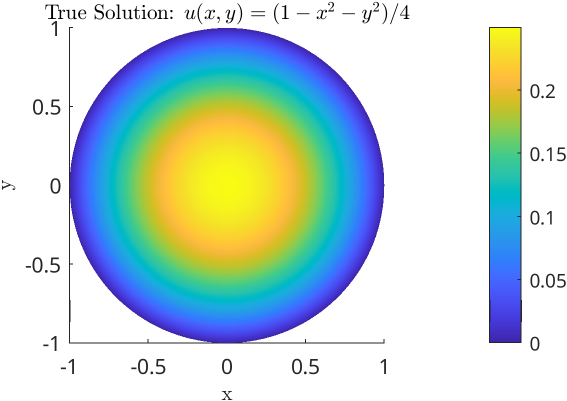}
      \caption{True solution}
      \label{fig:pred_both_ovfix}
    \end{subfigure}
    \caption{PIELM solution for Poisson's equation}\label{fig_poisson}
\end{figure}
%----------------------------------------------------------------------------------------------------------%
We refer to the Poisson equation test case published by MATLAB's official documentation\footnote{\url{https://fr.mathworks.com/help/pde/ug/solve-poisson-equation-on-unit-disk-using-pinn.html}}. We consider MATLAB's PINN solution as a reference and solve the same problem with PIELM.
\begin{itemize}
\item \textbf{Problem Description}: $-\left(\frac{\partial^{2}u}{\partial x^{2}}+\frac{\partial^{2}u}{\partial y^{2}}\right)=1$ in $\varOmega$, $u=0$ on $\delta\varOmega$, where $\varOmega$ is the unit disk. The exact solution is $u(x,y)=(1-x^{2}-y^{2})/4$
\item \textbf{Result}: Figure \ref{fig_poisson} presents the results obtained using PIELM for this problem, while Table \ref{Tab_PINNvsPIELM} provides a comparison with PINN. Similar to PINNs, there is no well defined  formula for selecting the optimal number of neurons in the hidden layer, as discussed in previous studies \citep{DWIVEDI202096,10.1115/1.4046892,DONG2022111290}. However, unlike traditional numerical methods, where truncation error can be systematically reduced, the error in PIELM solutions does not necessarily converge to zero by merely increasing the number of hidden-layer neurons. In this test case, we used fewer than half the number of parameters compared to PINN and only a small fraction of the data points. The results confirm that PIELM surpasses PINN in both speed and accuracy as a linear solver.
\begin{table}
\begin{centering}
\begin{tabular}{|c|c|c|}
\hline 
 & PINN & PIELM\tabularnewline
\hline 
\hline 
Data points & $13657$ & $1905$\tabularnewline
\hline 
Parameters & $5301$ & $2000$\tabularnewline
\hline 
Activation & $tanh$ & $tanh$\tabularnewline
\hline 
$L_{2}$Error & $\sim1e-1$ & $\sim1e-5$\tabularnewline
\hline 
Time taken  & $97$ s & $0.5$ s\tabularnewline
\hline 
\end{tabular}
\par\end{centering}
\caption{PINN vs PIELM for linear 2D Poisson's equation}
\label{Tab_PINNvsPIELM}
\end{table}
\end{itemize}
\FloatBarrier
%----------------------------------------------------------------------------------------------------------%
%   BURGERS 
%-------------------------------------------------------------------------------------------------------%
\subsection{Burgers' Equation} 
%-------------------------------------------------------------------------------------------------------%
\begin{figure}[ht]
 \centerline{\includegraphics[width=0.95\textwidth]{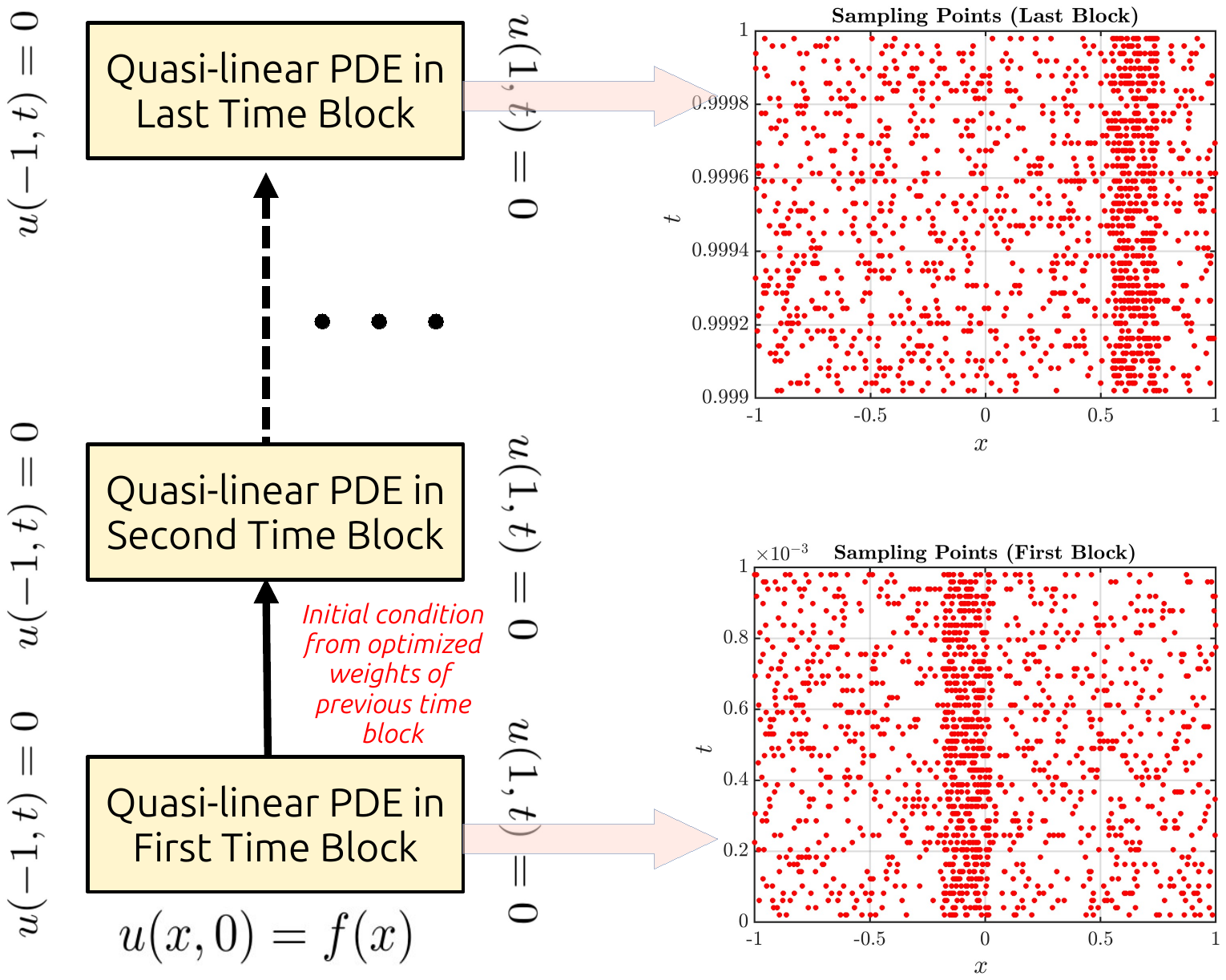}}
  \caption{(Left) Time-marching strategy for solving Burgers' equation. (Right) Example of RBF kernel distribution for the first and last time blocks, assuming a traveling wave initial condition \( f(x) = e^{-30x^2} \). Higher kernel density is used near regions with steep gradients in the initial condition of the time block.}  
\label{fig:Burgers+sampling}
\end{figure}
%-------------------------------------------------------------------------------------------------------%
\begin{figure}[ht]
    \centering
    \begin{subfigure}{0.5\textwidth}
      \centering
      \includegraphics[width=\textwidth]{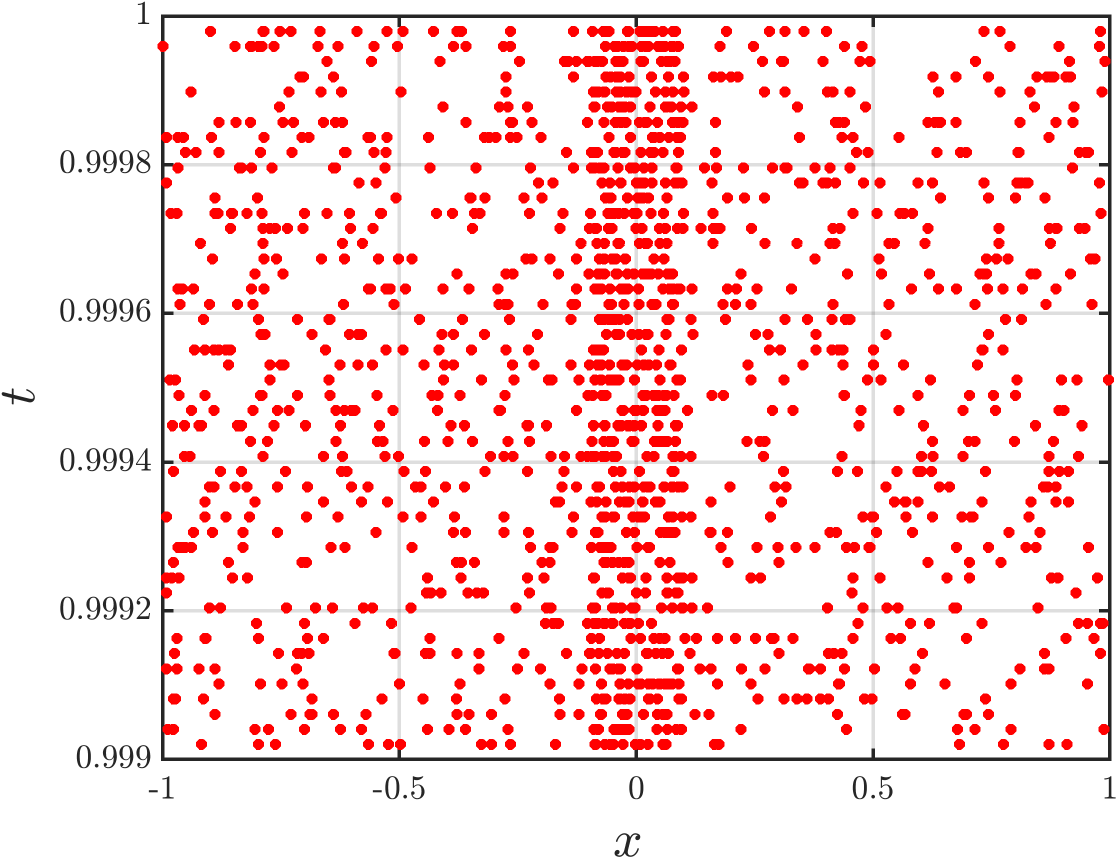}
      \caption{Spatial distribution of sampling points.}
      \label{fig:SAMP_POINTS_BUGERS_STANDING_SHOCK}
    \end{subfigure}
    \begin{subfigure}{0.5\textwidth}
      \centering
      \includegraphics[width=\textwidth]{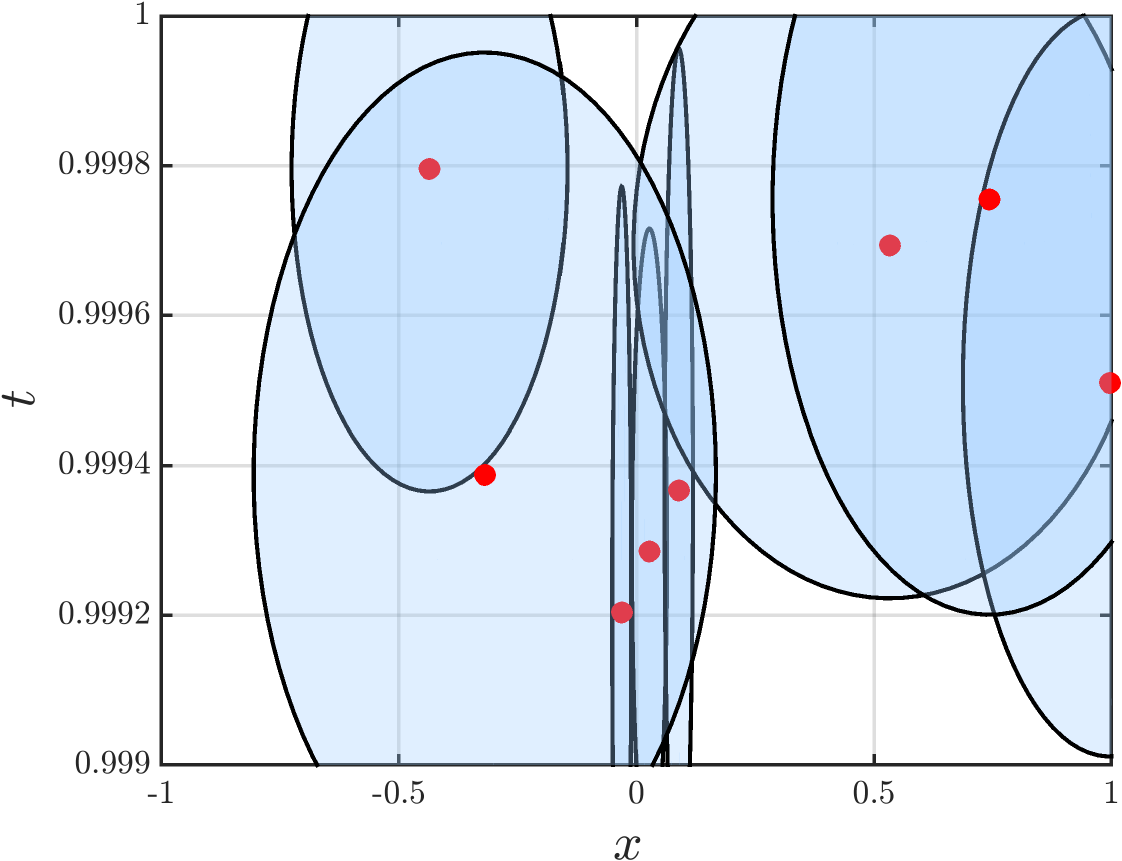}
      \caption{Spatial variation of RBF kernel standard deviations.}
      \label{fig:SIGMA_BUGERS_STANDING_SHOCK}
    \end{subfigure}
    \caption{Visualization of sampling points and the spatial variation of RBF kernel standard deviations.}
    \label{fig: BURGERS_SAMPLING_POINTS}
\end{figure}
%-------------------------------------------------------------------------------------------------------%
\begin{figure}[ht]
 \centerline{\includegraphics[width=0.98\textwidth]{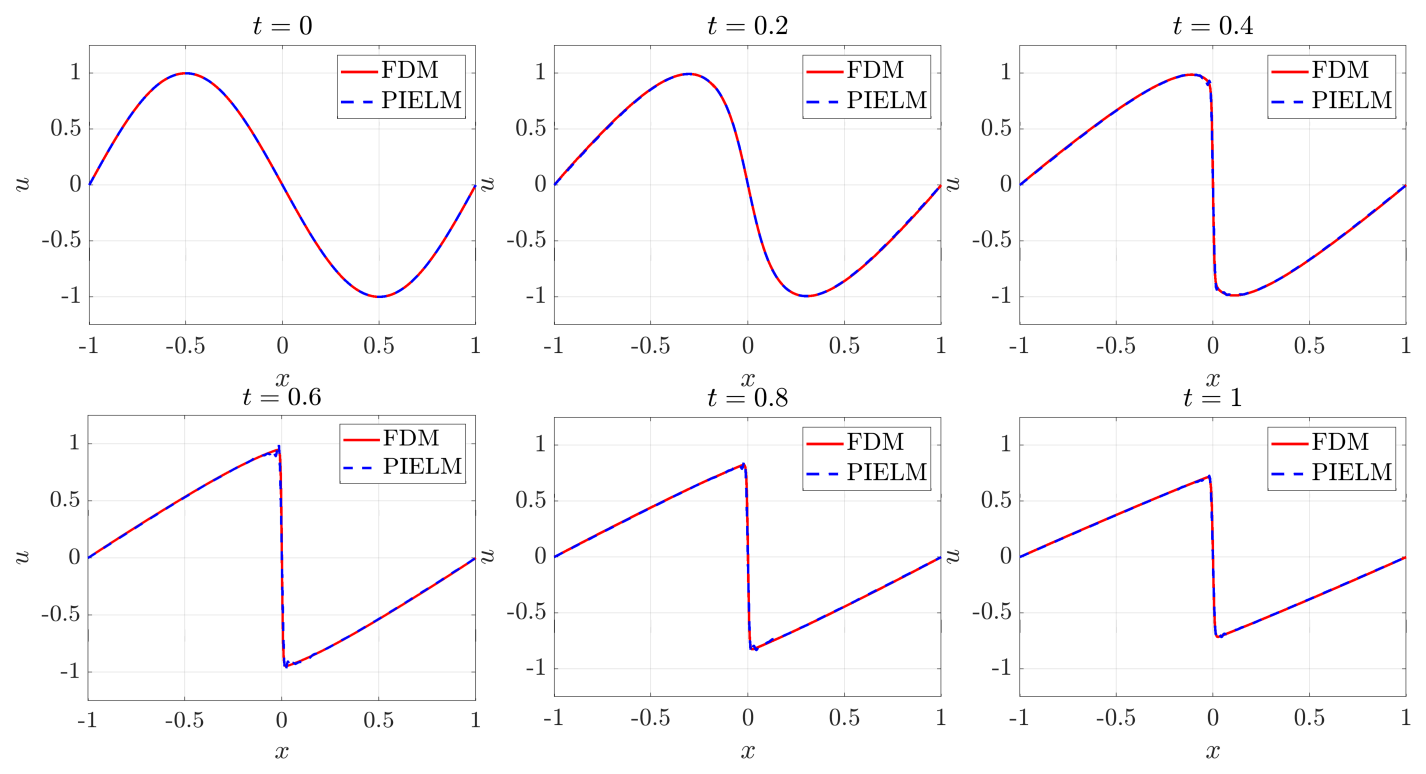}}
  \caption{Temporal evolution of the PIELM solution for the viscous Burgers' equation allowing a standing shock.}  
\label{fig:Standing_Shock}
\end{figure}
%-------------------------------------------------------------------------------------------------------%
\begin{figure}[ht]
 \centerline{\includegraphics[width=0.98\textwidth]{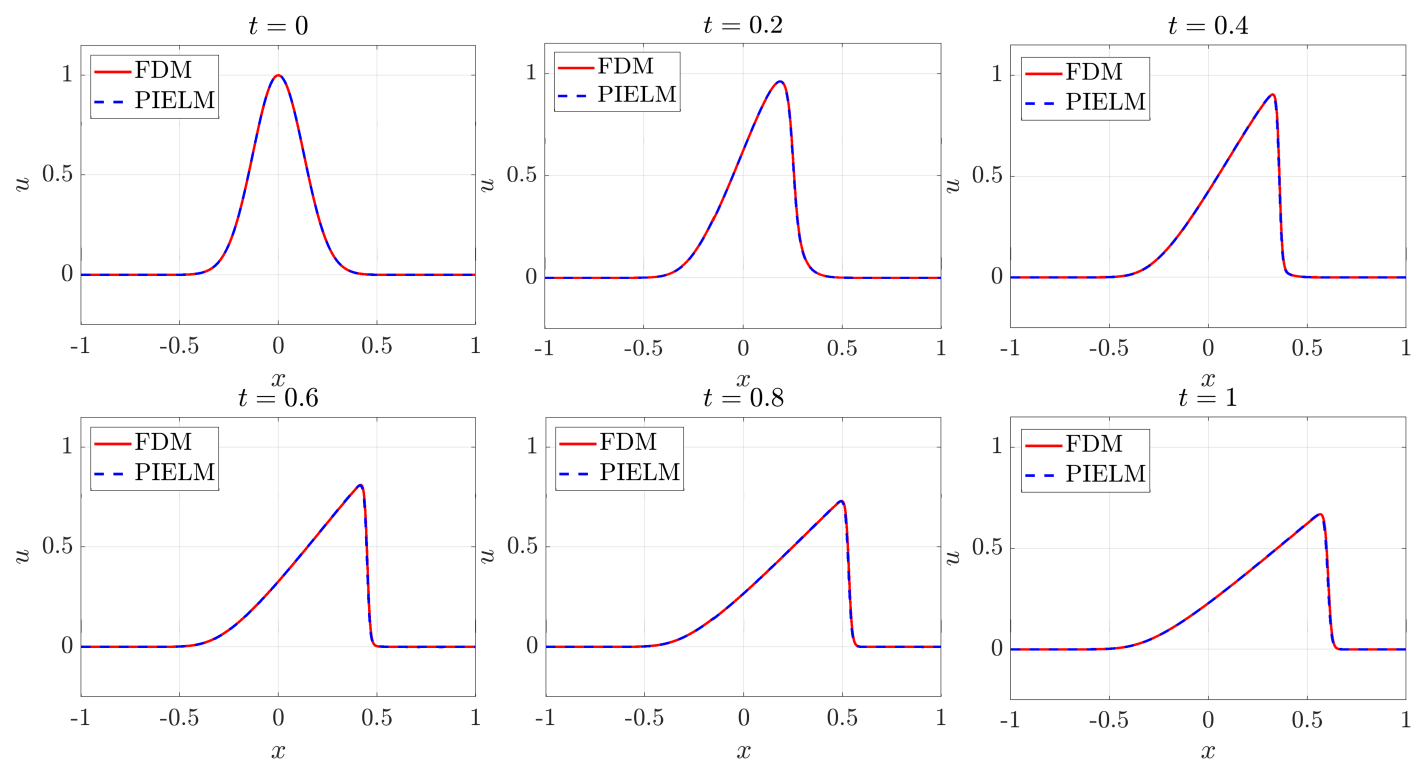}}
  \caption{Temporal evolution of the PIELM solution for the viscous Burgers' equation allowing a travelling shock.}  
\label{fig:Traveling_Shock}
\end{figure}
%-------------------------------------------------------------------------------------------------------%

We solve Burgers' equation \ref{eqn:Burgers} over the computational domain \([-1,1] \times [0,1]\) with a small viscosity parameter \(\nu = 0.01/\pi\). The Dirichlet boundary conditions are given by \(u(-1,t) = 0\) and \(u(1,t) = 0\). We consider two types of initial conditions:  
\begin{enumerate}  
    \item A Gaussian hump, which evolves into a traveling shock solution.  
    \item A sine wave, which results in a standing shock solution.  
\end{enumerate}  

PIELM employs a time-marching strategy to solve Burgers' equation, where the optimal weights from the previous time block initialize the current block. The gradient of the initial condition is computed to identify regions with steep variations. To effectively capture shocks, a small traveling window is defined, and the absolute sum of gradients within this window is evaluated. The window with the highest gradient sum is identified, and sampling points are concentrated in this region to enhance resolution. Additionally, the standard deviation of RBF kernels in this high-gradient region is kept smaller compared to the rest of the computational domain, improving local approximation accuracy.

In this study, the traveling window size is set to 0.1, with 30$\%$ of the total collocation points allocated within this region. For the standing shock case, the high-gradient region remains fixed at the center, whereas for the traveling shock case, the shock position continuously shifts. The sampling point distribution for the traveling shock case is illustrated in Figure \ref{fig:Burgers+sampling}. The figure demonstrates that a higher density of RBF kernels is placed in regions with steep initial condition gradients. Figure \ref{fig: BURGERS_SAMPLING_POINTS} provides a visualization of the spatial variation of the standard deviations of the RBF kernel, showing that the spread of the RBF kernels in high-gradient regions is narrower than in other areas. Mathematically, 
\[
\sigma_x^{(j)} =
\begin{cases}
\text{Uniform}(0.01, 0.04), & \text{if } x_{L} \leq \alpha^{*}_j \leq x_{R}, \\
\text{Uniform}(0.02, 0.6), & \text{otherwise}.
\end{cases}
\]

\noindent Here, \( x_{L} \) and \( x_{R} \) define the high-gradient region, and "uniform" refers to a uniform random distribution. The standard deviations inside and outside the shock region are chosen based on corresponding length scales. For \( \sigma_t \), relatively larger values (0.4-0.6 times the time step size) are selected. It should be noted that the physical interpretation of the PIELM parameters enables the selection of their ranges based on both physical and geometrical aspects of the problem.
  
The details of the PIELM solution for the standing and traveling shock cases are summarized in Table \ref{Tab:Burgers_Cases} and the results are shown in Figures \ref{fig:Standing_Shock} and \ref{fig:Traveling_Shock}. PIELM successfully captures both types of shock solutions, demonstrating its effectiveness. For comparison with PINN, we refer to the official documentation of MATLAB\footnote{\url{https://fr.mathworks.com/help/deeplearning/ug/solve-partial-differential-equations-with-lbfgs-method-and-deep-learning.html}}, which presents a PINN-based solution for the viscous shock test case. The PINN implementation utilizes a second-order optimizer and the high-performance \texttt{dlaccelerate}\footnote{\url{https://fr.mathworks.com/help/deeplearning/ref/dlaccelerate.html}} training routine, achieving accurate results with 3000 parameters in 15 minutes. In particular, PIELM produces comparable results with only 1500 learning parameters, highlighting its computational efficiency.

\begin{table}
\begin{centering}
\begin{tabular}{|c|c|c|}
\hline 
 & Travelling wave & Standing wave\tabularnewline
\hline 
\hline 
Initial condition & $f(x)=e^{-30x^{2}}$ & $f(x)=-sin(\pi x)$\tabularnewline
\hline 
Viscosity  & $0.01/\pi$ & $0.01/\pi$\tabularnewline
\hline 
Data points per block & $2113$ & $2113$\tabularnewline
\hline 
Parameters & $1500$ & $1500$\tabularnewline
\hline 
Activation & RBF & RBF\tabularnewline
\hline 
Time blocks & $1000$ & $1000$\tabularnewline
\hline 
Time taken & $10.7$ mins & $13$ mins\tabularnewline
\hline 
\end{tabular}
\par\end{centering}
\caption{PIELM solution details for Burgers' equation test cases.}
\label{Tab:Burgers_Cases}
\end{table}
%--------------------------------------------------------------------------------------------------%
\subsection*{Remarks}
\begin{itemize}
\item In weakly nonlinear cases (i.e., for large values of $\nu$), identifying high-gradient regions and concentrating sharp RBF kernels is not required. However, for shock cases, this step is essential; otherwise, the PIELM solution exhibits non-physical oscillations (see Figure \ref{fig:Oscillations} ), regardless of the number of sampling points or RBF kernels.
%-------------------------------------------------------------------------------------------------------%
\begin{figure}[ht]
 \centerline{\includegraphics[width=0.98\textwidth]{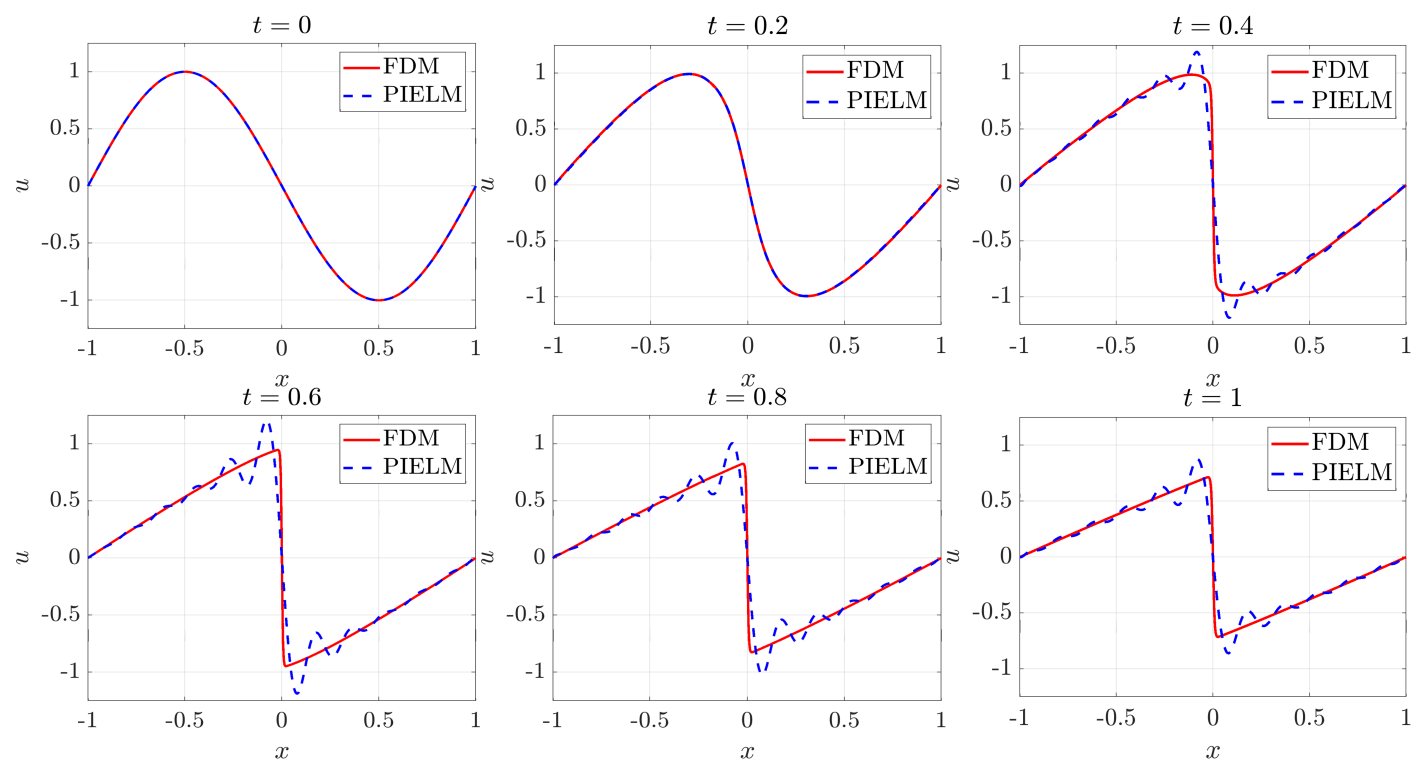}}
  \caption{Unphysical oscillations in the PIELM solution for standing shock case.}  
\label{fig:Oscillations}
\end{figure}
%-------------------------------------------------------------------------------------------------------%

\item Ideally, the size of the steep variation or shock region should be determined by the order of gradients and should vary with time. Likewise, the spread of the RBF kernels should also be time-dependent. However, in this study, the shock region width is kept fixed at 0.1 throughout the simulation. Consequently, we observe that PIELM performs better after the shock has fully developed compared to the early stages of its formation. 

\item In conventional time-marching methods, the time step size is governed by the spatial grid resolution through the CFL condition\citep{10.5555/2430727}. In our approach, effective shock resolution using PIELM requires a sufficient number of RBF centers and a carefully selected time step size. The precise mathematical criterion for determining these parameters remains an open question. 

\item Beyond spatio-temporal resolution, the stability of the PIELM solution is also influenced by the shape of the RBF kernels. If $\sigma_{t}$ is too small in comparison to the size of the time step, the solution becomes unstable and diverges.    

\end{itemize}

\FloatBarrier
%----------------------------------------------------------------------------------------------------------%
%   NAVIER STOKES
%----------------------------------------------------------------------------------------------------------%
\subsection{Lid-Driven Cavity Flow}
%----------------------------------------------------------------------------------------------------------%
\begin{figure}[ht]
    \centering
    \begin{subfigure}{0.45\textwidth}
      \centering
      \includegraphics[width=\textwidth]{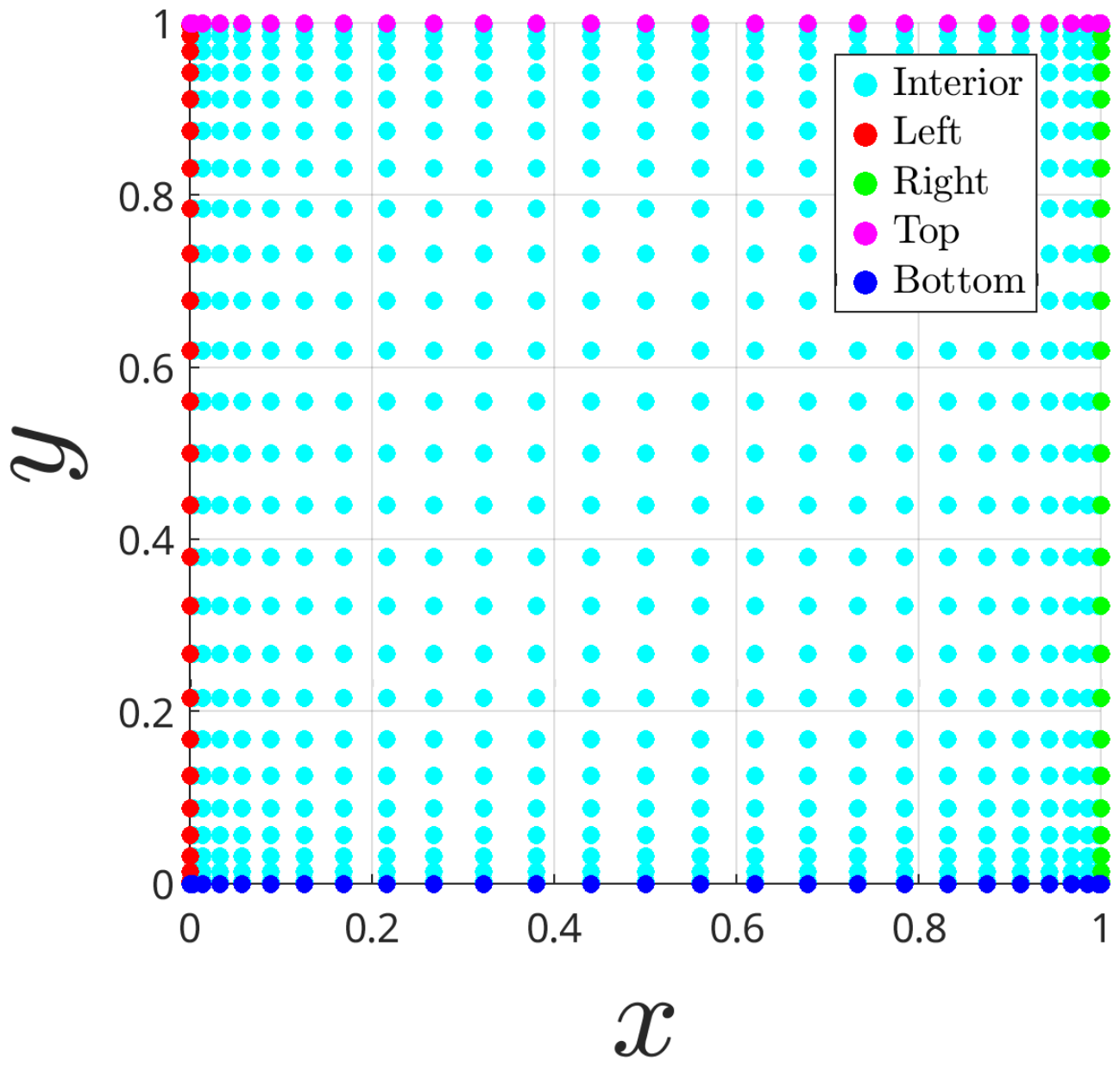}
      \caption{Spatial distribution of sampling points.}
      \label{fig:LDC_SAMP_POINTS}
    \end{subfigure}
    \begin{subfigure}{0.55\textwidth}
      \centering
      \includegraphics[width=\textwidth]{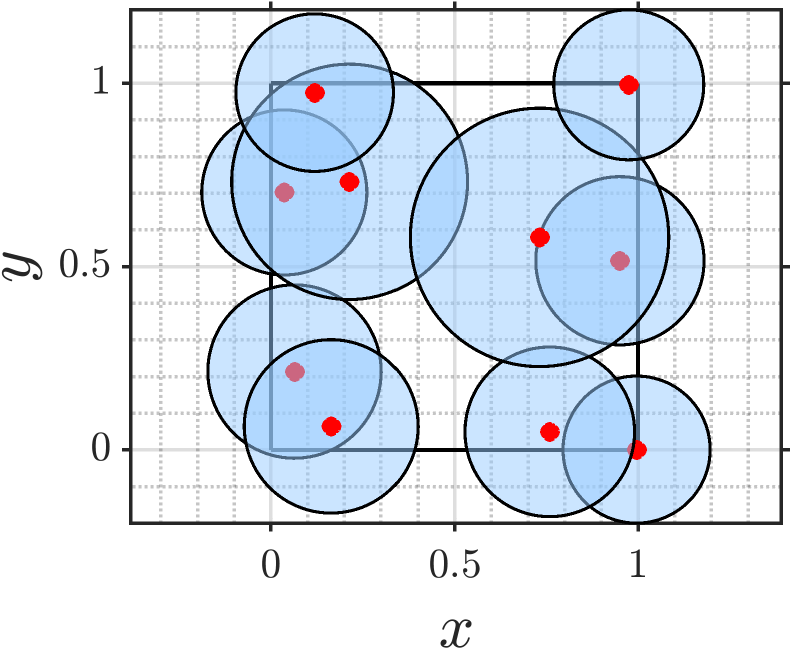}
      \caption{Spatial variation of RBF kernel standard deviations.}
      \label{fig:LDC_SIGMA}
    \end{subfigure}
    \caption{Sampling points distribution and RBF-kernel width variation for the lid-driven cavity flow.}
    \label{fig: LDC_geometry}
\end{figure}
%----------------------------------------------------------------------------------------------------------%
\begin{figure}[ht]
 \centerline{\includegraphics[width=0.99\textwidth]{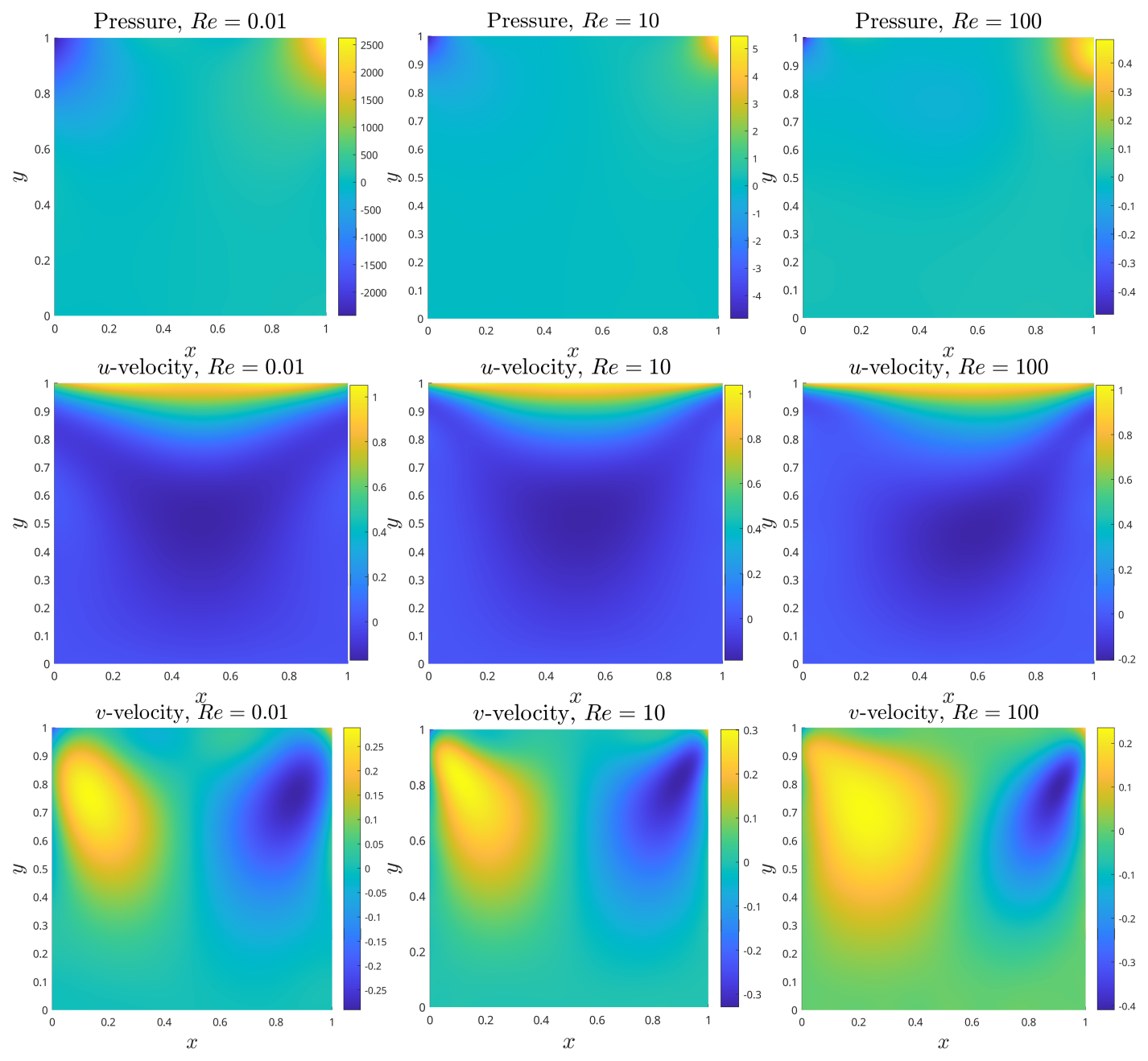}}
  \caption{PIELM predicted velocity components and pressure for lid-driven cavity flow.}
  \label{fig_uvp_PIELM}
\end{figure}
%----------------------------------------------------------------------------------------------------------%
\begin{figure}[ht]
 \centerline{\includegraphics[width=0.95\textwidth]{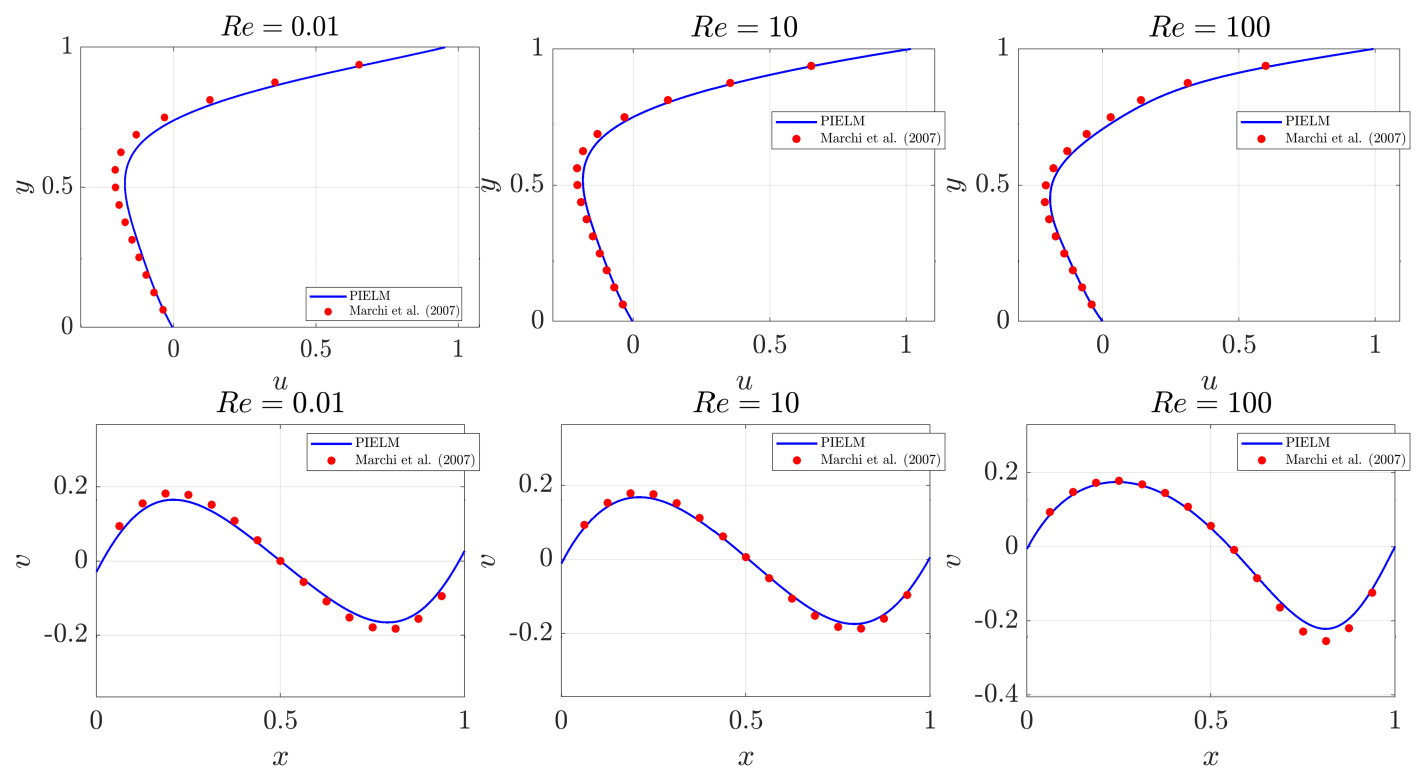}}
  \caption{Comparison of centerline velocities with high resolution finite volume simulations\citep{marchi2009lid} for lid-driven cavity flow.}\label{fig_centerline_PIELM}
\end{figure}
%----------------------------------------------------------------------------------------------------------%
\begin{figure}[ht]
 \centerline{\includegraphics[width=0.99\textwidth]{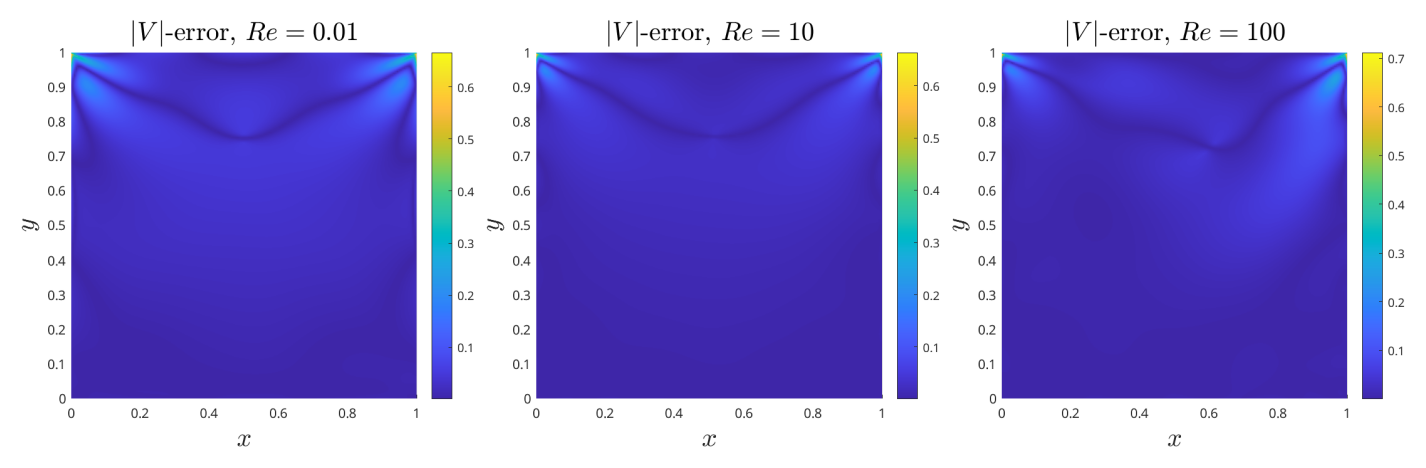}}
  \caption{Error distribution of velocity magnitude at $Re = 0.01, 10, 100$}
  \label{fig_error_dist_at_Re_10_100}
\end{figure}
%----------------------------------------------------------------------------------------------------------%
\begin{figure}[ht]
    \centering
    \begin{subfigure}{0.45\textwidth}
      \centering
      \includegraphics[width=\textwidth]{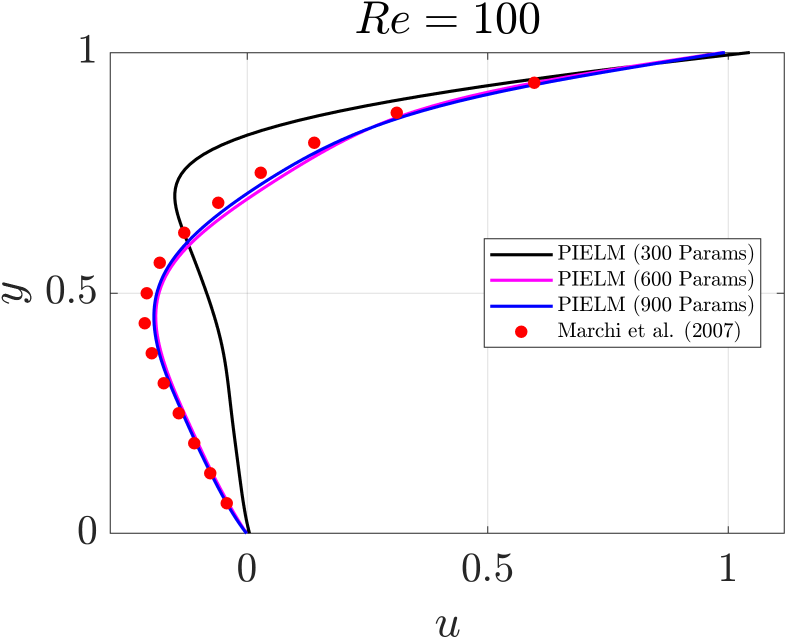}
      \caption{Effect on vertical velocity component.}
      \label{fig:NN_ver}
    \end{subfigure}
    \begin{subfigure}{0.45\textwidth}
      \centering
      \includegraphics[width=\textwidth]{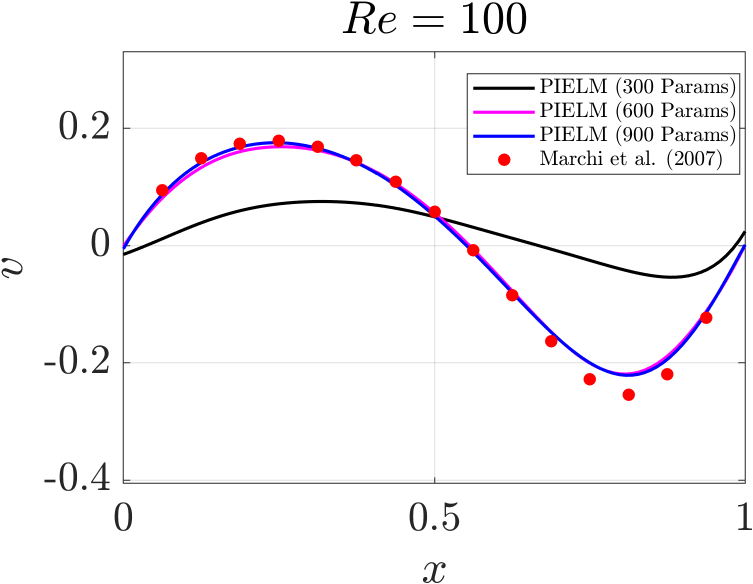}
      \caption{Effect on horizontal velocity component.}
      \label{fig:NN_hor}
    \end{subfigure}
    \caption{Effect of number of PIELM parmeters on centerline velocity prediction.}
    \label{fig: NN_effect}
\end{figure}
%----------------------------------------------------------------------------------------------------------%
In this section, we solve the classic lid-driven cavity problem\citep{marchi2009lid}. The lid-driven cavity problem is a well-known benchmark in fluid dynamics for testing numerical methods that solve the incompressible Navier-Stokes equations. It consists of a unit-square cavity ($[0,1]$ \texttimes{} $[0,1]$) filled with a viscous, incompressible fluid, where the top lid moves at a constant horizontal velocity while the other three walls remain stationary. A no-slip boundary condition is applied, ensuring zero velocity at the walls except for the moving lid, which drives the flow inside the cavity. The fluid is assumed to be of unit density, and therefore the Reynolds number is given by $Re=1/\nu$. 

Mathematically, the governing system of PDEs are Navier-Stokes equations\ref{eqn:cont},\ref{eqn:mom} and the boundary conditions are expressed as follows:
\begin{itemize}
\item At the left edge, $\overrightarrow{V}(0,y)=(0,0)$
\item At the bottom edge, $\overrightarrow{V}(x,0)=(0,0)$
\item At the right edge, $\overrightarrow{V}(1,y)=(0,0)$
\item At the top edge, $\overrightarrow{V}(x,1)=(1,0)$
\item At the bottom left corner, $p(0,0)=0$
\end{itemize}

Figure \ref{fig: LDC_geometry} depicts the distribution of sampling points and spatial variation of RBF kernel widths. A total of \( 25 \times 25 \) sampling points are used, arranged with Chebyshev spacing to concentrate more points near the walls for capturing sharp gradients. For PIELM, only 300 RBF kernels have been employed, as the lid-driven cavity flow remains laminar for Reynolds numbers up to 100. Similar to the sampling points, a denser placement of RBF kernels is maintained near the walls. 

The standard deviation (\(\sigma\)) of the RBFs is determined based on their proximity to the boundaries, following the relation \(\sigma = 0.2 + 0.4 (L_{\min} / L_{\max})\), where \(L_{\min}\) denotes the minimum distance of an RBF center from the boundary walls, and \(L_{\max}\) is the maximum possible distance. For a unit square domain, \(L_{\max} = 0.7071\). Compared to a random selection of $\sigma$, our choice of PIELM parameters is intuitive and entirely based on physical aspects of the problem. 

The lid-driven cavity flow problem has previously been solved by PIELM using the vorticity-streamfunction approach for Stokes flow\citep{10.1115/1.4046892}. To our knowledge, this is the first attempt to solve it with PIELM in terms of $(u,v,p)$ up to a Reynolds number of 100. Figure \ref{fig_uvp_PIELM} presents the PIELM solutions for velocity components and pressure at \(Re = 0.01, 10, 100\). To evaluate the accuracy of PIELM, we compare our results with an established reference. Since an exact analytical solution for the lid-driven cavity problem is unavailable, we use high-resolution finite-volume simulations from Marchi et al. \citep{marchi2009lid} as a benchmark. 

Figure \ref{fig_centerline_PIELM} illustrates the velocity profiles along the horizontal and vertical centerlines of the square, comparing the PIELM predictions with the well-established results of Marchi et al. \citep{marchi2009lid} at \(Re = 0.01, 10, 100\). The comparison confirms that PIELM accurately captures the centerline velocity distributions, demonstrating its effectiveness in solving the lid-driven cavity problem. Figure\ref{fig: NN_effect} shows the effect of number of PIELM parameters on the results. The error in PIELM solutions decreases as we increase the number of neurons from 300 to 600 but this drop saturates and does not converge to zero by merely increasing the number of hidden-layer neurons.

To further analyze the error distribution across the domain, we compare the PIELM solution with the finite element-based solver FEniCS \citep{LoggWells2010} on a high-resolution (\(250 \times 250\)) mesh. The FEM implementation used for this comparison is available online\footnote{\url{https://github.com/cpraveen/fenics/blob/master/2d/ns_cavity/demo_steady.py}}. 

The root mean square errors (RMSE) for the velocity magnitude are 0.039 at \(Re = 0.01\), 0.026 at \(Re = 10\) and 0.034 at \(Re = 100\), with the corresponding error distributions shown in Figure \ref{fig_error_dist_at_Re_10_100}. As expected, the highest errors occur near the top-left and top-right corners, where velocity discontinuities are present. 

At first glance, it may seem counterintuitive that the RMSE at \(Re = 100\) is lower than at \(Re = 0.01\). However, as discussed in the second remark of the previous section, this behavior arises because the standard deviation of the RBF kernels is not adaptive and remains fixed across different Reynolds numbers. 

For comparison with PINNs, we refer to a recent advanced variant called SPINN \citep{RAMABATHIRAN2021110600}. A comparison with SPINN reveals the computational efficiency of PIELM. For the lid-driven cavity problem at \(Re = 100\), PIELM demonstrated superior performance, utilizing only 900 learnable parameters and completing the computation in 130 seconds. In contrast, SPINN required approximately 820 seconds to solve the same problem. This significant difference in computation time highlights PIELM's substantial speed advantage over PINN-based solvers.

For the lid-driven cavity problem up to $Re=100$, a Reynolds number step size of 0.1 was used. However, applying the same step size up to $Re=400$ results in inaccuracies in the PIELM solution. At this higher Reynolds number, the flow becomes advection-dominated, forming two secondary vortices near the bottom corners. Capturing these features requires more data points, additional neurons, and a reduced Reynolds number step size. However, determining the optimal values for these parameters remains an open question.
\FloatBarrier
%-----------------------------------------------------------------------------------------------------------%
\subsection{Stenotic Flow}
Stenosis refers to the narrowing of a blood vessel, resulting in an increase in blood velocity in the stenotic throat due to the reduced cross-sectional area. Understanding blood flow in stenotic vessels is crucial in cardiovascular research, as it plays an important role in the development and progression of various cardiovascular diseases. Computational fluid dynamics (CFD) simulations provide a detailed analysis of flow patterns under different physiological conditions, such as rest and exercise. These simulations can also capture the effects of pulsatile flow and complex stenotic geometries.

\begin{figure}[ht]
    \centering
    \begin{subfigure}{0.8\textwidth}
      \centering
      \includegraphics[width=\textwidth]{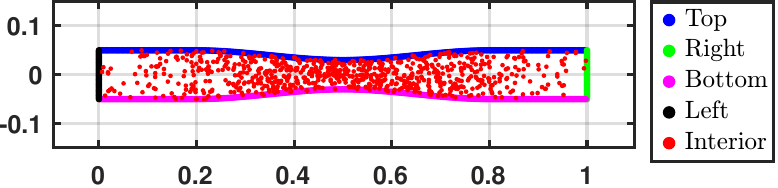}
      \caption{Spatial distribution of sampling points.}
      \label{fig:SAMP_POINTS_STENOSIS}
    \end{subfigure}
    \begin{subfigure}{0.8\textwidth}
      \centering
      \includegraphics[width=\textwidth]{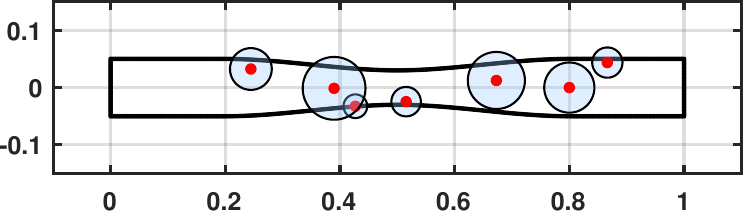}
      \caption{Spatial variation of RBF kernel standard deviations.}
      \label{fig:SIGMA_STENOSIS}
    \end{subfigure}
    \caption{Sampling points distribution and RBF-kernel width variation for the stenotic flow.}
    \label{fig: Stent_geometry}
\end{figure}

In this section, we test PIELM to model blood flow in a stenotic vessel by solving the steady-state Navier-Stokes equations \ref{eqn:cont} and \ref{eqn:mom}. Figure \ref{fig: Stent_geometry} depicts the computational geometry along with the distribution of sampling points. It also shows the spatial variation of RBF-kernel widths. The stenotic vessel has a total length of 1 unit (\(L=1\)), with an inlet width of 0.1 unit (\(R_{in}=0.1\)) and a constriction width of 0.06 unit (\(D_{c}=0.06\)). The geometry is symmetric, with both the inlet and outlet extending 0.2 units.  

To effectively capture flow dynamics, sampling points are distributed exponentially, ensuring a higher density near the throat. The placement of RBF centers follows the same principle. The standard deviation ($\sigma$) of the RBFs is determined based on their proximity to the boundaries, following the relation $\sigma=min(D_{c},0.02+D_{c}(L_{min}/D_{c}))$, where $L_{min}$ denotes the minimum distance of an RBF center from the boundary walls. For this study, it is assumed that the fluid has a unit density (\(\rho=1\)).

The boundary conditions are defined as follows. At the inlet, the velocity follows a parabolic profile:  

\begin{equation}
\overrightarrow{V}=\left(\begin{array}{c}
u\\
v
\end{array}\right)=\left(\begin{array}{c}
U_{max}\left(1-\left(\frac{y}{R_{in}}\right)^{2}\right)\\
0
\end{array}\right)
\end{equation}

where $U_{max}=0.1$. No-slip conditions are imposed on the top and bottom walls, enforcing 

\begin{equation}
\overrightarrow{V} = \left( \begin{array}{c} 0 \\ 0 \end{array} \right).
\end{equation}

At the outlet, the pressure is set to zero (\(p = 0\)), and symmetric velocity boundary conditions are applied:

\begin{equation}
\frac{\partial u}{\partial x} = 0, \quad \frac{\partial v}{\partial x} = 0.
\end{equation}

The Reynolds number is determined using the equation: 
\begin{equation}
Re=\frac{\rho U_{max}R_{in}}{\mu}
\end{equation} 
To control the Reynolds number, we vary its value by adjusting the dynamic viscosity $\mu$. 

\begin{figure}[ht]
 \centerline{\includegraphics[width=0.95\textwidth]{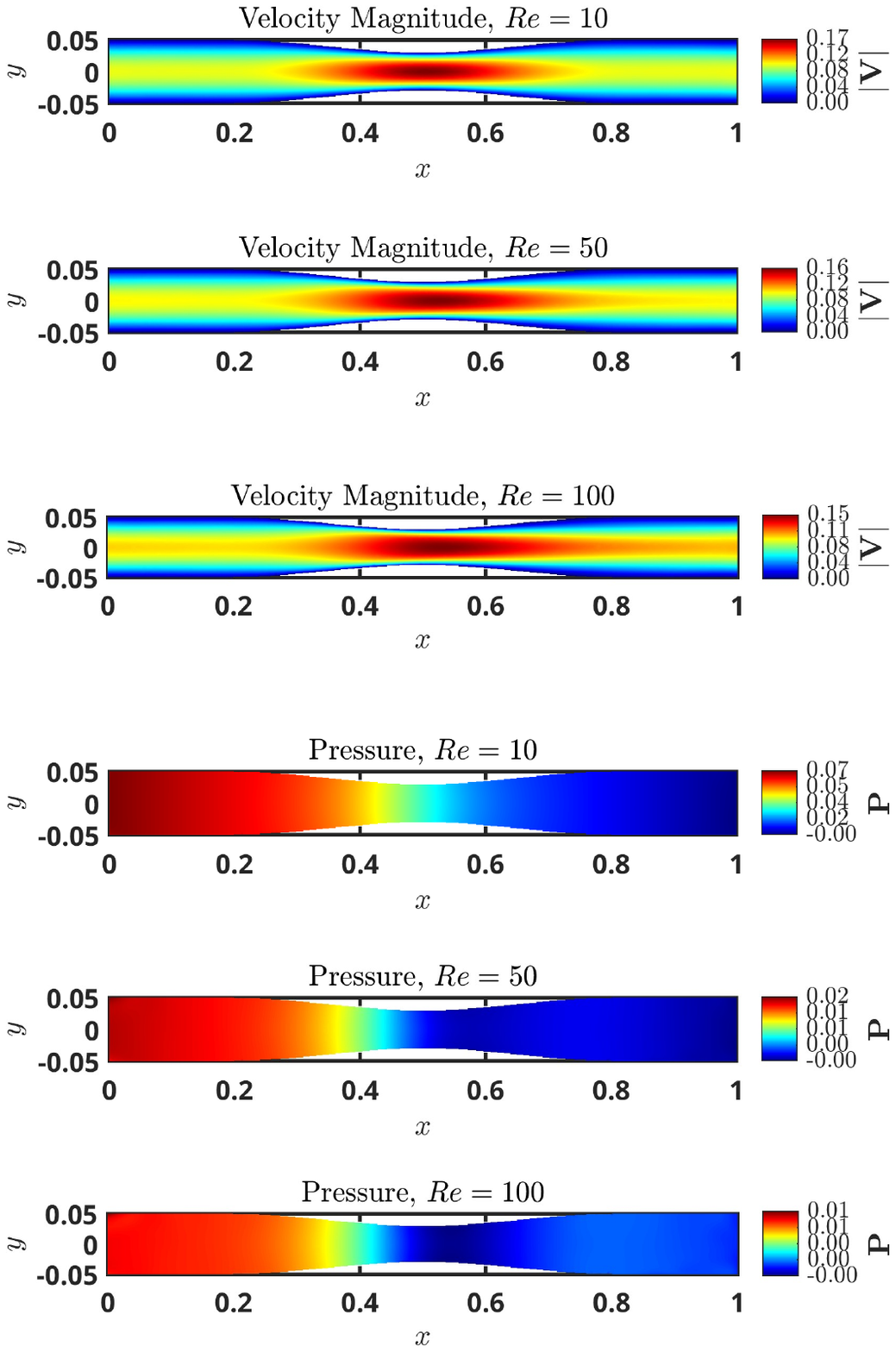}}
  \caption{PIELM predicted solutions for the stenotic flow at $Re=10, 50$ and $100$.}
  \label{fig: pielm_stokes_stent}
\end{figure}

\begin{figure}[ht]
 \centerline{\includegraphics[width=0.85\textwidth]{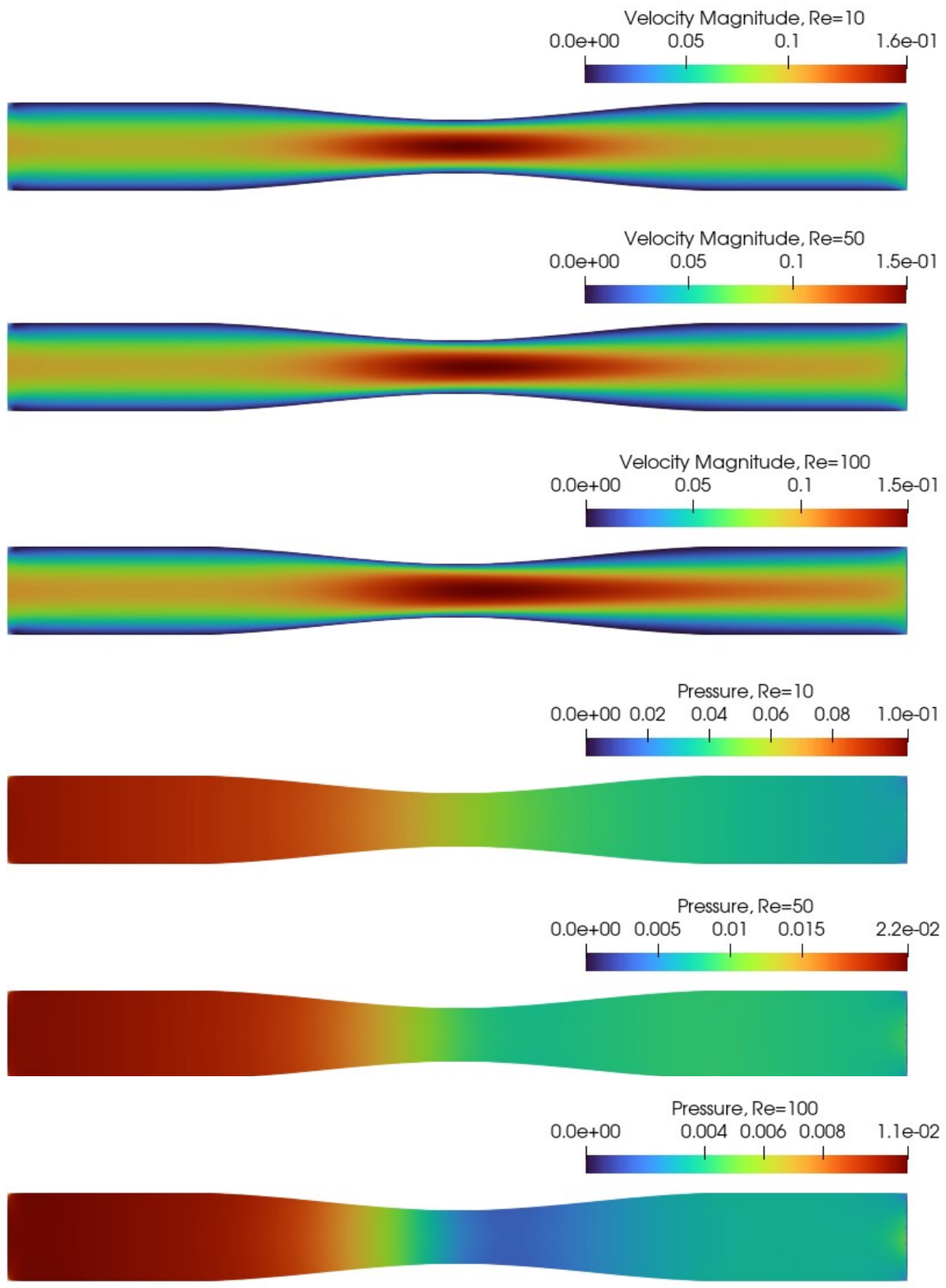}}
  \caption{FEM solutions for the stenotic flow at $Re=10, 50$ and $100$.}
  \label{fig: fem_stokes_stent}
\end{figure}

Figure \ref{fig: pielm_stokes_stent} presents the PIELM-predicted solutions for stenotic flow at \( Re = 10, 50, \) and \( 100 \), while the corresponding FEM results are shown in Figure \ref{fig: fem_stokes_stent}. The PIELM simulation employs 1000 PDE sampling points, 310 boundary points, and 2400 trainable parameters, whereas the FEM solution is computed using 13,959 total points and 27,124 triangular elements. The PIELM method accurately captures essential flow features, particularly the high velocity at the stenotic throat. The root mean square errors (RMSEs) between the PIELM and FEM solutions for velocity magnitude and pressure at \( Re = 10, 50, \) and \( 100 \) are (0.0053, 0.0267), (0.0054, 0.0071), and (0.005, 0.003), respectively. As discussed in earlier sections, the counter-intuitive decrease in error with increasing $Re$ is attributed to the fixed range of RBF spread across different values of $Re$.
%----------------------------------------------------------------------------------------------------------%

\FloatBarrier
%-----------------------------------------------------------------------------------------------------% 
%                                      Conclusions                                                    %
%-----------------------------------------------------------------------------------------------------%
\section{Conclusion}\label{Sec:Conclusion}

In this paper, we presented two physics-informed machine learning algorithms for solving nonlinear PDEs related to fluid flow by combining PIELM, curriculum learning, and RBF-kernels. While PIELMs are significantly faster than PINNs for linear PDEs, their direct application to nonlinear PDEs remains challenging. To overcome this, we integrated PIELMs with curriculum learning to solve nonlinear problems by iteratively solving their quasilinear approximations, tackling nonlinearity in a gradual and controlled manner. Building on the SPINN framework, we employed RBF kernels as activation functions, which establish a direct connection between PIELM function approximation and the RBF kernel collocation method, improving interpretability. We assessed the performance of curriculum-driven PIELMs on benchmark cases, including the Burgers' and Navier-Stokes equations, and analyzed their strengths and limitations. We also employed PIELM to simulate stenotic flow as a practical application. Based on our findings, we highlight the following key conclusions:
\begin{enumerate}
\item Interpretable PIELMs guided by curriculum learning were formulated to solve both steady and unsteady nonlinear PDEs.
\item A clear physical meaning is associated with the input layer parameters of the proposed PIELMs, corresponding to the centers and standard deviations of RBF kernels.
\item To the best of our knowledge, this study is the first in which PIELM has been applied to:
\begin{itemize}
\item Solve Burgers' equation for traveling and standing shock cases.
\item Solve Navier-Stokes equations up to $Re=100$ using a primary variable formulation.
\end{itemize}
\item The proposed PIELMs demonstrate a significant speed advantage over PINNs when solving weakly nonlinear PDEs. For example, while SPINN, a state-of-the-art PINN variant, takes 13 minutes to solve the lid-driven cavity flow at $Re=100$, PIELM completes the same task in just 2 minutes.
\item The proposed PIELMs have been found to be comparable to PINNs in solving strongly nonlinear PDEs in terms of speed, while being more efficient in terms of the number of learnable parameters.
\item Unlike typical optimization algorithms where the training history lacks physical meaning, in this case, the iterations correspond to fluid flow at gradually increasing Reynolds numbers.
\item Beyond the benchmark cases, the curriculum learning-driven PIELM is employed to simulate stenotic flow as a practical application. 
\end{enumerate}

The proposed PIELM framework has potential applications in various areas, including microfluidics (Stokes flow, creeping flow), aerodynamics of small drones (low Reynolds number flows), incompressible internal flows (pipe and channel flows), groundwater and porous media flows, ocean and atmospheric flows (tidal currents, climate modeling), and biofluid dynamics (blood flow in large arteries). Future research will focus on expanding applications in these areas, and on automating PIELM hyperparameter selection.
%----------------------------------------------------------------------------------------------------------%
%-----------------------------------------------------------------------------------------------------%
%\section*{Acknowledgments}
%This was was supported in part by......

%Bibliography
%\bibliographystyle{unsrt} 
\bibliographystyle{apalike}
\bibliography{references}

\end{document}